\documentclass[letterpaper]{article} % DO NOT CHANGE THIS
\usepackage{aaai2026}  % DO NOT CHANGE THIS
\usepackage{times}  % DO NOT CHANGE THIS
\usepackage{helvet}  % DO NOT CHANGE THIS
\usepackage{courier}  % DO NOT CHANGE THIS
\usepackage[hyphens]{url}  % DO NOT CHANGE THIS
\usepackage{graphicx} % DO NOT CHANGE THIS
\usepackage{natbib}  % DO NOT CHANGE THIS AND DO NOT ADD ANY OPTIONS TO IT
\usepackage{caption} % DO NOT CHANGE THIS AND DO NOT ADD ANY OPTIONS TO IT
\usepackage{algorithm}
\usepackage{algpseudocode}   % 提供 algorithmicx 的 pseudocode 样式
\usepackage{amsmath} % ok?
\usepackage[most]{tcolorbox}

\usepackage{makecell}   % ok?
\usepackage{graphicx}   % ok?
\usepackage{subcaption} % 创建子图
\usepackage{multirow}  % compute cost table
\usepackage{xcolor} % 加载 xcolor 宏包
\usepackage{caption}  % 用于非浮动图片的标题和编号
\usepackage{subcaption} % 用于子图标题
\usepackage{enumitem} % 用于自定义列表样式
\usepackage[utf8]{inputenc} % allow utf-8 input

\usepackage{url}            % simple URL typesetting
\usepackage{booktabs}       % professional-quality tables
\usepackage{amsfonts}       % blackboard math symbols
\usepackage{nicefrac}       % compact symbols for 1/2, etc.
\usepackage{microtype}      % microtypography
\usepackage{xcolor}         % colors

\usepackage{newfloat}
\usepackage{listings}
\DeclareCaptionStyle{ruled}{labelfont=normalfont,labelsep=colon,strut=off} % DO NOT CHANGE THIS
\floatstyle{ruled}
\newfloat{listing}{tb}{lst}{}
\floatname{listing}{Listing}
\title{PromptFlow: Training Prompts Like Neural Networks}
\author {
    Jingyi Wang,
    Hongyuan Zhu,
    Ye Niu,
    Yunhui Deng
}
\affiliations {
    Alibaba Cloud\\
    wangjingyi.w@alibaba-inc.com,
    zhy19933628@gmail.com,\\
    niuye@alibaba-inc.com,
    yunhui.dyh@alibaba-inc.com
}
\usepackage{bibentry}
\begin{document}

\maketitle

\begin{abstract}
Large Language Models (LLMs) have demonstrated profound impact on Natural Language Processing (NLP) tasks. However, their effective deployment across diverse domains often require domain-specific adaptation strategies, as generic models may underperform when faced with specialized data distributions. Recent advances in prompt engineering (PE) offer a promising alternative to extensive retraining by refining input instructions to align LLM outputs with task objectives. This paradigm has emerged as a rapid and versatile approach for model fine-tuning. Despite its potential, manual prompt design remains labor-intensive and heavily depends on specialized expertise, often requiring iterative human effort to achieve optimal formulations. To address this limitation, automated prompt engineering methodologies have been developed to systematically generate task-specific prompts. However, current implementations predominantly employ static update rules and lack mechanisms for dynamic strategy selection, resulting in suboptimal adaptation to varying NLP task requirements. Furthermore, most methods treat and update the whole prompts at each step, without considering editing prompt sections at a finer granularity. At last, in particular, the problem of how to recycle experience in LLM is still underexplored. To this end, we propose the PromptFlow, a modular training framework inspired by TensorFlow, which integrates meta-prompts, operators, optimization, and evaluator. Our framework can be equipped with the latest optimization methods and autonomously explores optimal prompt refinement trajectories through gradient-based meta-learning, requiring minimal task-specific training data. Specifically, we devise a reinforcement learning method to recycle experience for LLM in the PE process. Finally, we conduct extensive experiments on various datasets, and demonstrate the effectiveness of PromptFlow.
\end{abstract}


\section{Introduction}
Large language models (LLMs) have been emerging and achieving state-of-the-art (SOTA) results on a variety of natural language processing tasks. This success has propelled prompt engineering to become a hot topic. Current approaches predominantly rely on manually crafted prompts incorporating domain-specific knowledge. For instance, \citet{wei2021finetuned} proposed instruction tuning through natural language template development, improving zero-shot performance on multiple NLP benchmarks. For complex reasoning tasks, \citet{wei2021finetuned} introduced chain-of-thought (CoT) prompting, gaining accuracy in mathematical reasoning through stepwise decomposition mechanisms. Currently, multi-perspective optimization frameworks \citep{wang2022self, shinn2023reflexion, yao2024tree} emerged, employing self-refinement and tree-search algorithms to enhance task-specific adaptability. Simultaneously, there are exciting systems and tools \citep{tenney2024interactive,diao2023active}, aiming to help engineers develop prompts efficiently. For example, \citet{tenney2024interactive} presented a visual tool for interactive prompt debugging with input salience methods. However, while these methods can enhance performance on specific tasks through the creation of extensive prompts, they also require substantial manpower and domain expertise.

\begin{figure*}[htbp]
    \centering
    \includegraphics[width=0.8\textwidth]{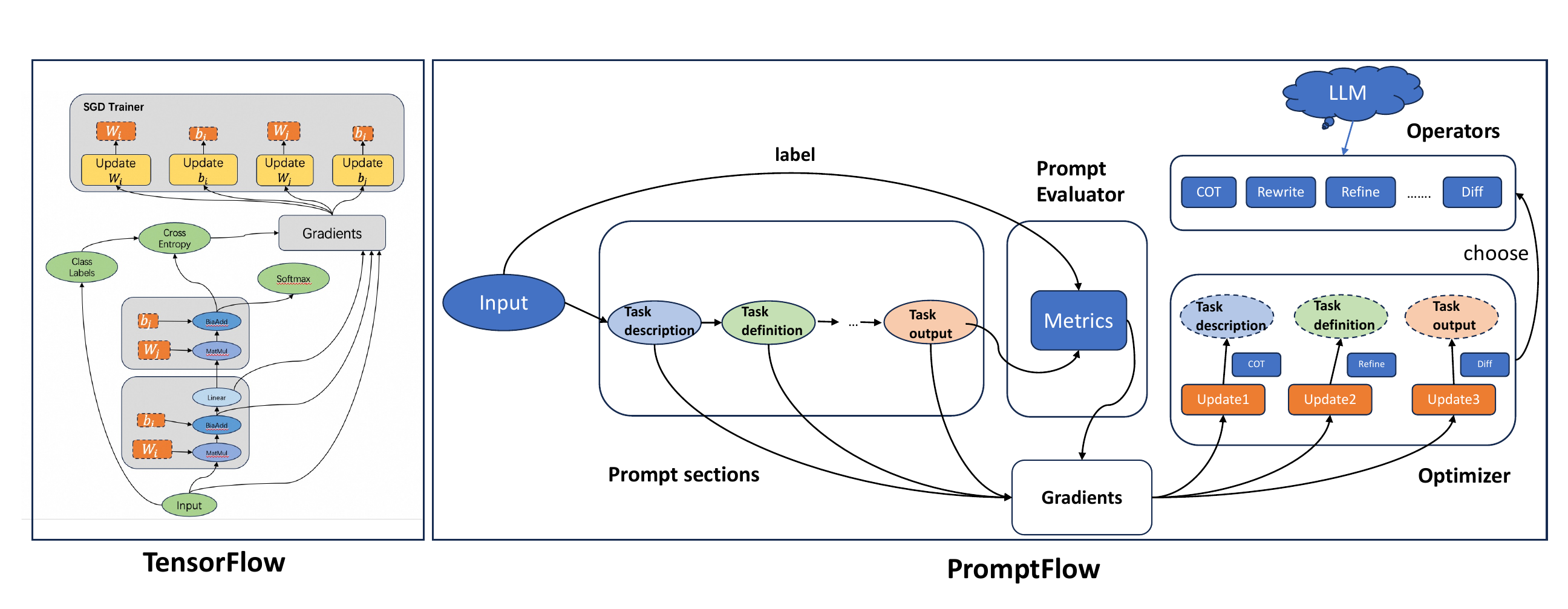}
    \caption{Comparison of TensorFlow and PromptFlow}
    \label{fig:promptflow}
\end{figure*}

To enhance the efficiency and effectiveness of prompt engineering, multiple studies \citep{zhou2022large,schnabel2024prompts,hsieh2023automatic} explored automated prompt generation frameworks. \citet{zhou2022large} pioneered a methodology that leverages large language models to create instruction pools followed by candidate selection mechanisms. These approaches represent a significant advancement by enabling automatic prompt generation through template rewriting. Recent advancements focus on optimizing iterative update strategies. \citet{liu2023large} developed an evolutionary algorithm-based framework demonstrating superior performance and rapid convergence in prompt generation. In parallel, \citet{ye2023prompt} introduced meta-prompt architectures incorporating hierarchical components, such as two-stage task decomposition, contextual specifications, and structured reasoning templates, to enhance LLM-driven prompt synthesis. ~\citet{tang2024unleashing,pryzant2023automatic} emphasized the strategic optimization of update mechanisms to improve both performance metrics and convergence rates. Notably, \citet{tang2024unleashing} innovatively adapted gradient-based optimization principles to design enhanced update strategies for LLM-based prompt optimizers, marking a significant methodological advancement in the field.

Indeed, automatic prompt generation still faces several challenges that have yet to be fully addressed by existing approaches. In particular, current methods rely on predefined prompt refinement strategies, such as rewriting, COT reasoning, and self-correction. However, the development of adaptive mechanisms capable of dynamically adjusting and transitioning update strategies for diverse scenarios remains insufficient. Furthermore, existing experience-based feedback strategies easily overhaul entire prompts during refinement, risking the degradation of high-quality sections. A more effective paradigm would involve selectively refining underperforming segments of prompts while retaining their effective components. As prior works employ fixed methods to address specific NLP tasks, they often demonstrate limited extensibility to novel scenarios or emerging optimization techniques, thereby necessitating task-specific redesigns.

To address these limitations, we introduce PromptFlow, an end-to-end framework for prompt generation and training that effectively handles a wide range of diverse NLP tasks. Inspired by TensorFlow, PromptFlow integrates meta-prompts, operators, optimizers, and evaluators. Specifically, we propose a meta-prompt generator, which can generate prompt sections by utilizing various operators, such as self-reflection, COT, differential evolution, and so on. We then employ a meta-level and gradient-inspired optimization mechanism that integrates reinforcement learning to intelligently select operators and determine the most effective refinement paths for specific prompt segments. In this way, PromptFlow can identify suitable operators and discover optimized paths tailored to different tasks. We conduct comprehensive experiments across multiple benchmarks on three datasets, including named entity recognition (NER), classification (CLS), and machine reading comprehension (MRC). The results demonstrate that PromptFlow can significantly enhance prompt performance and demonstrate superiority over baselines on massive NLP tasks. The contributions of this paper are summarized as follows:

\begin{itemize}
    \item We propose the PromptFlow, a modular training framework. PromptFlow enables fine-grained improvements of prompts at the meta-prompt level, dynamically assembles prompt components using a rich and extensible library of operators, including self-reflection, chain-of-thought reasoning, differential evolution and so on.
    \item We develop a meta-level and gradient-based optimization mechanism that intelligently selects operators and identifies optimal refinement paths for specific prompt segments. Furthermore, we incorporate a reinforcement learning (RL) method, which effectively recycles and leverages experience to enhance the performance of large language models.
    \item We conduct comprehensive experiments across multiple benchmarks, covering three datasets that span diverse tasks such as named entity recognition, classification, and machine reading comprehension. The results indicate that PromptFlow can effectively enhance prompt performance and consistently outperform baseline methods.
\end{itemize}

\section{Related Work}

\subsection{Prompt Engineering}

A prompt in generative model is the textual input provided by users to guide the model's output. Prompting offers a natural and intuitive interface for humans to interact with LLMs. Existing work has designed methods with human knowledge effectively to improve the model’s performance on specific tasks. For instance, \citet{wei2021finetuned} used an instruction tuning method to improve zero-shot~\cite{palatucci2009zero} performance on unseen tasks by developing natural language instruction templates. To improve performance in complex reasoning, \citet{wei2022chain} proposed chain-of-thought prompting to reason and solve problems step by step.

However, designing effective prompts manually is both challenging and time-consuming. To handle this, several researches \citep{zhou2022large,schnabel2024prompts,hsieh2023automatic} have proposed methods for automatically generating prompts. \citet{zhou2022large} proposed a method that leverages LLMs to generate a pool of prompt candidates and subsequently select the most suitable candidates. These methods are capable of generating new prompts automatically, but their generation mechanisms are predominantly template-based or rule-based, which limits the adaptability and flexibility of the generation process. Recent research mainly focuses on rewriting the entire prompt. However, for complex NLP tasks,  prompts can be more intricate, and modifying the entire prompt may significantly impact the results while risking the loss of effective components from the original prompt.

\subsection{Large Language Model}

In recent years, global enthusiasm for large language models (LLMs) has surged dramatically and has demonstrated remarkable performance across a wide range of natural language tasks. GPT-4, a massive transformer-based model developed by OpenAI \citep{achiam2023gpt}, has proven its ability to match human performance in a multitude of professional and academic assessments. Following its success, GPT-4o (omni)\citep{hurst2024gpt}, the upgraded iteration, has swiftly been introduced, significantly expanding the frontiers of capability and performance. At the same time, Meta’s Llama 3\citep{grattafiori2024llama},  which supports multilingual tasks, coding, reasoning, and tool integration, has demonstrated performance on par with top-tier models such as GPT-4 across a variety of practical applications. The Qwen series\citep{yang2024qwen2} incorporates both dense models and a Mixture-of-Experts framework to deliver advanced capabilities. These have achieved markedly improved performance in downstream applications, particularly in solving intricate and demanding problems.

\subsection{Reinforcement Learning}

Reinforcement learning (RL) is a computational framework for developing decision-making policies through iterative interactions with dynamic environments. In previous studies, reinforcement learning has been successfully applied across applications of machine learning, from natural language processing \citep{yu2017seqgan,ouyang2022training} to computer vision \citep{mirowski2016learning,silver2016mastering}. For instance, \citet{silver2016mastering} integrated value networks for position evaluation and policy networks for strategic planning, enabling the Monte Carlo tree search to master gameplay. More recently, reinforcement learning methods has taken a central role in enhancing large language models (LLMs). The seminal work by \citet{ouyang2022training} established a paradigm for human-aligned LLM training through reinforcement learning from human feedback (RLHF), 
utilizing preference rankings to fine-tune GPT-3.\citet{deng2022rlprompt} developed  RLPROMPT, which constructed a policy network by training a task-specific multilayer perceptron (MLP) module and its parameters were updated based on reward signals generated through RL methods. Subsequent studies have expanded RL’s applicability in LLMs.\citet{achiam2023gpt} systematically analyzed safety alignment via RLHF, while \citet{rafailov2023direct} developed direct preference optimization methods that circumvent explicit reward modeling.\citet{kwon2024stableprompt} proposed StablePrompt, in which the target LLM combined with a given dataset served as the world model, while the agent LLM acted as the policy and the reward was derived from the target LLM's response. Notably, \citet{guo2025deepseek} demonstrated that large-scale reinforcement learning (RL) training, even without supervised fine-tuning initialization, can significantly enhance reasoning capabilities.

\section{Problem Statement}

A common scene of prompt engineering is writing prompt for LLMs to solve a problem based on specific datasets and requirements. Thus, for a NLP task, we define a dataset $D$, considering as a pair of input and output $\left \{ X,Y \right \} $. $X$ usually contains the task description and context, and $Y$ is defined as the ground truth output of task. Our objective is to generate the correct output $y$ by combining prompt $P$ and $X$ as the input of LLM. $P = \left \{ p_{1}, p_{2},\dots, p_{l_{p}} \right \} $ is the generated prompt, and $p_{i}$ is the i-th section of the prompt. In this paper, we discuss a strategy $G_{\beta }$ to generate best prompt efficiently and accurately. The $G_{\beta }$ employs operators and optimizers to explore additional possibilities while leveraging feedback from the evaluator $E$ for self-learning. The goal of prompt optimization is to find the optimal prompt $p^{*}$ drawn from the natural language space that maximizes the expectation of the score over $D$. ${M}_{T}$ denotes the LLM used, and $F$ refers to the evaluation metrics. Formally, the problem of prompt optimization can be formulated as:

\begin{equation}
p^{*} = \underset{p\sim \mathcal{M}_{T} }{argmax} E_{\left \langle x,y \right \rangle\in D } \left [ F\left ( \mathcal{M}_{T}\left ( x;G_{\beta\left (p  \right )  } \right ), y  \right )  \right ]
\end{equation}

\section{PromptFlow} \label{sec:Prompt Generation Pipeline}

\begin{figure*}[h]
    \centering
    \includegraphics[width=0.9\linewidth]{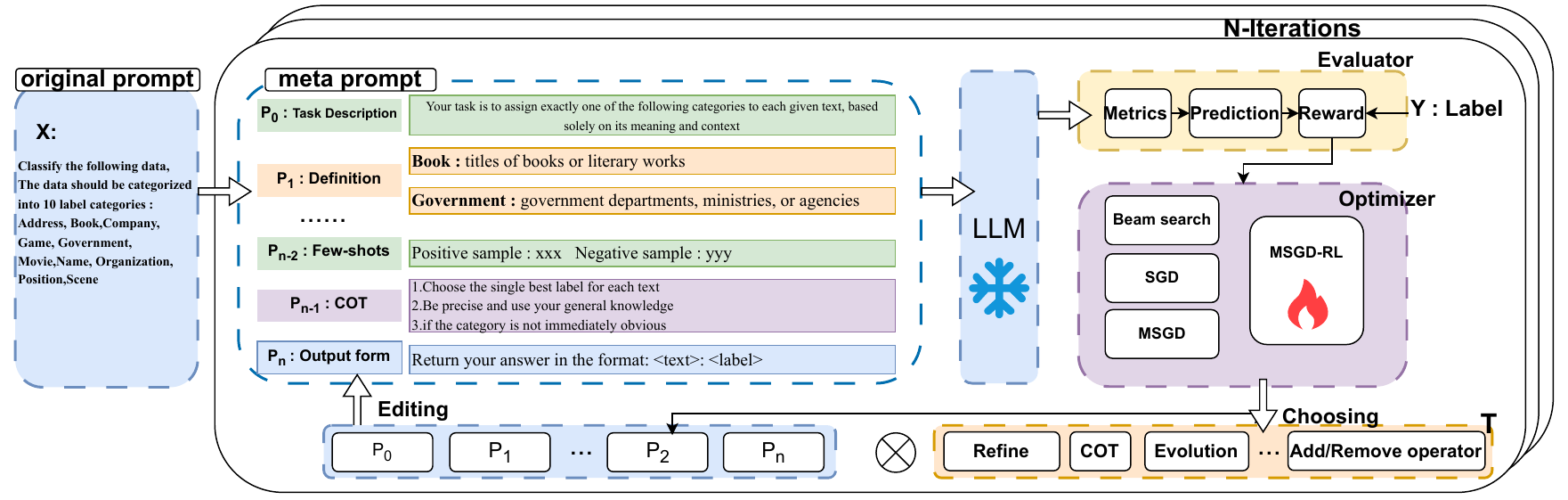}
    \caption{Prompt training and generation pipeline}
    \label{fig:pipeline}
\end{figure*}

The PromptFlow framework comprises four core components: Meta-Prompt, Operator, Optimizer, and Evaluator. Inspired by~\citet{developers2022tensorflow}'s groundbreaking approach to training and inference in neural networks, we propose that LLMs similarly necessitate a specialized framework for the systematic optimization of prompts across diverse domains. Drawing a parallel to TensorFlow’s methodology, which leverages GPU acceleration to train predefined architectures through parameter optimization from massive datasets, PromptFlow establishes a structured approach for prompt engineering. Similar like fine-tuning model parameters by minimizing loss using gradient-based optimizers, PromptFlow adopts similar core principles for prompt evolution. PromptFlow uses LLMs together with training data annotated with ground truth labels to optimize prompts, incorporating various operators such as COT, Few-Shot, Self-Reflection and more. Based on evaluation metrics and an evaluator, PromptFlow receives feedback from optimizer by calculating the loss between predictions and ground truth. Figure~\ref{fig:pipeline} shows the pipeline of prompt generation.

\subsection{Meta-Prompt}
Existing approaches typically rewrite entire prompts during refinement, risking degradation of high-quality sections. A more effective paradigm would selectively refine underperforming prompt segments while preserving effective components. Therefore, we define meta-prompt as a sequence $P = \left \{p_{0}, p_{1},\dots ,p_{m}\right \}$, where $p_{i}$ represents the $i$-th section of $P$. As shown in Figure~\ref{fig:pipeline}, there are various meta-prompts, such as task description, definition, few-shots, and output format, which together constitute a complete prompt. In our framework, meta-prompts can be generated either from scratch or starting from an initial template.

\subsection{Operator}

We have systematically collected and developed a series of methodological enhancements for prompt engineering, referred to as Operators. These techniques, such as COT, Few-Shot, and Self-Reflection, have been empirically shown to significantly enhance the performance of LLMs. The design of Operators emphasizes extensibility, allowing newly validated techniques to be seamlessly integrated into the existing framework without additional development effort. Due to space constraints, here we elaborate on a subset of most frequently used  Operators below.

\textbf{COT} enables model to break down multi-step problems into intermediate steps, enhancing their reasoning abilities to handle complex problems~\citep{wei2022chain}.

\textbf{Self-Reflection} is a powerful approach used in model reasoning, which could enhance its performance by reflecting on previous experience~\citep{shinn2023reflexion}.

\textbf{Differential Evolution} utilizes individual differences among existing prompts to generate new candidates and progressively improves prompt quality through candidate selection~\citep{guo2023connecting}.

\textbf{Define Sort} The sequencing of meta-prompts may influence the final outcome, as sentences closer to the output tend to exert a greater effect. Therefore, we introduce this operator to reorder meta-prompts accordingly.

\textbf{Merge} is applied when multiple high-performing prompts exist. It merges the best components of each to achieve globally optimal results. 

\textbf{Few-Shot} Previous study \citep{brown2020language} compares one-shot and few-shot methods with the zero-shot method and observes that one-shot and few-shot always have higher performance.We utilize few-shot operators to extract samples from origin dataset utilizing different sampling strategies to help achieve higher performance.

\textbf{Short-Instruction} is employed to streamline content or make it more concise, thereby enhancing the performance of large language models (LLMs) by minimizing the disruption caused by irrelevant information. 

\textbf{Self-Consistency} generates multiple possible meta-prompts and selects the most consistent and reasonable one, effectively avoiding errors that may arise from relying on a single solution ~\cite{wang2022self}.

\textbf{Repeat Instructions}. We employ repeat instructions to reinforce key points in the meta-prompts, ensuring sustained focus on the specific task. 

\textbf{RAG} is applied when working on knowledge-based problems. By utilizing Rag, we can search for useful knowledge, incorporating prompt to gain better result~\cite{riedel2020retrieval}.

\subsection{Optimizer}

Existing research~\citep{zhou2022large} utilized a random sampling method to generate new prompts as next epoch candidate prompts, which would cost massive time and computing power for each optimization. Inspired by gradient descent, ~\citet{pryzant2023automatic} and ~\citet{tang2024unleashing} proposed a gradient-based approach to sample better prompts, which consider the loss between predictions and ground truth, utilizing the loss to feedback and refine prompts. These works could cleverly sample and choose better prompts, avoiding a violent exploration of all possibilities. However, they refine the entire prompt without considering the possibility of optimizing the prompt section by section. It is more effective to refine only the suboptimal components of prompts while preserving the high-quality content, as this minimizes unnecessary disruptions and leverages strengths of the existing prompt. By focusing on targeted modifications, the optimization process becomes more efficient and maintains the integrity of well-performing segments, ultimately leading to better overall performance. In this way, we propose a meta-level stochastic gradient descent (MSGD) Optimizer to handle this problem, which could calculate the gradient of each part of prompts separately and update sections of prompts to bring positive returns.

\subsubsection{MSGD Optimizer}
Given meta-prompts $S$ and Operators $O$, MSGD optimizer is defined as a specific unit that focuses on selecting which operators to optimize prompts. We define the prompt $S = \{s_0, s_1, ..., s_l\}$ and operators $O = \{o_0, o_1, ..., o_n\}$. We utilize batch dataset $D=\{(x_0, y_0),(x_1, y_1), ..., (x_m, y_m)\}$ to train and refine prompt sections over $N$ epochs. In every epoch, the MSGD Optimizer decide to choose $o_j$ to rewrite $i$-th section of prompt $S$ based on value $Q(s_i, o_j)$,

\begin{equation}
\label{eq:q_matrix1}
Q_{ij} = \frac{exp(E_{S}^{i}{E_{O}^{j}}^{T} )}{ {\textstyle \sum_{i=0}^{i=l}\sum_{j=0}^{j=n}}exp(E_{S}^{i}{E_{O}^{j}}^{T} )}
\end{equation}

\begin{equation}
\label{eq:q_matrix2}
Q_{ij}^{t} = Q_{ij}^{t-1} + \alpha Norm(L(G(x, p^{*}_{t}), y)-L(G(x, p^{*}_{t-1}), y))
\end{equation}
where vectors $E_{S}^{i}$ and $E_{O}^{j}$ correspond to the selection probabilities of $i$-th section of prompt and $j$-th operator. A task-specific loss function $L$ measures the discrepancy between the model’s prediction and the ground truth. Here, $Norm$ denotes the normalization of the loss, and $\alpha$ is a hyperparameter. The errors are then embedded into the prompt as feedback, guiding the $G$ to edit new prompt $p^{*}$ in a direction that counteracts the semantic drift introduced by the prediction error, thus preparing it for the next iteration.

\subsubsection{MSGD-RL Optimizer}

Although the MSGD optimizer can refine prompts effectively through meta-learning, achieve rapid convergence, and reduce resource consumption, a significant challenge still remains: LLMs lack the ability to reuse prior experience when fine-tuning. When training on different datasets, existing methods typically require prompts to be trained from scratch. However, we notice that when humans optimize prompts to solve problems, they would draw upon past experiences and refine their strategies iteratively. This experiential learning allows for more efficient and effective prompt design, avoiding the need to start from scratch with each new data. To address this issue, we propose MSGD-RL Optimizer to learn and recycle experience in training processes, as shown in Algorithm~\ref{alg:MSGD-RL Optimizer}.

\begin{algorithm}[h]
\caption{MSGD-RL Optimizer}
\label{alg:MSGD-RL Optimizer}
\begin{algorithmic}[1]
\Require Large language model \textit{LLM}, evaluator \textit{E}, prompt generator $G$, dataset $D$, meta-prompts input $S = \{s_0, s_1, \ldots, s_n\}$, operators union $O = \{o_0, o_1, \ldots, o_l\}$, and transition matrix $M = [m_{00}, m_{01}, \ldots, m_{nl}]$
\Ensure $S' = \{s_0', s_1', \ldots, s_n'\}$, where $s_i'$ means the $i$-th section of refined prompt
\State Initialize $G$, $M$ with random weights $\theta$, $\phi$
\State Run $LLM$ using $S$ on $D$ and using $E$ to get loss $L$
\State Generate new prompts $S'$ using $O$ based on matrix $M$
\State Choose best-$k$ prompts for next epoch and update $M$ with $L$
\Repeat
    \For{g-steps do}
        \State Generate multiple sequences $S^* \sim G$
        \For{$t$ in $1:T$ do}
            \State Compute $Q(s = s_t; o = o_t)$ by Eq.~\ref{eq:q_max_matrix}
        \EndFor
        \State Update matrix parameters by Eq.~\ref{eq:update}
    \EndFor
    \For{d-steps do}
        \State Use best-$k$ prompts for next epoch and $a$ prompts as simulated annealing examples
        \State Evaluate new $S'$ for next epoch
    \EndFor
\Until{$L$ converges}
\end{algorithmic}
\end{algorithm}

The procession of using operator to refine the section of prompt could be defined as Markov Decision Process (MDP), as shown in Equation~\ref{eq:qq_max_matrix}.

\begin{equation}
\label{eq:qq_max_matrix}
Q^{k}_{\text{max}} =
\begin{bmatrix}
Q(s_0, o_0) & Q(s_0, o_1) & \cdots & Q(s_0, o_m) \\
Q(s_1, o_0) & Q(s_1, o_1) & \cdots & Q(s_1, o_m) \\
\vdots & \vdots & \ddots & \vdots \\
Q(s_n, o_0) & Q(s_n, o_1) & \cdots & Q(s_n, o_m)
\end{bmatrix}
\end{equation}

The State $S$ is the current prompt. The action $A$ is to choose operator $a_t$ to rewrite prompt section $s_t$. The reward $R$ is determined by evaluating the predicted result against the ground truth. The $S^{\prime}$ is the new prompt after rewriting. The $A^{\prime}$ is next action to refine the prompt. Based on the track $\left ( S, A, R, S^{\prime }, A^{\prime}   \right)$ and inspired by Sarsa method, we use $Q(s_{t+1}, a_{t+1})$ to update $Q(s_t, a_t)$.

% \begin{equation}
% \label{eq:this update}
% \begin{split}
% Q(s_t, a_t) \leftarrow Q(s_t, a_t) + 
% \alpha \left( r_{t+1} + \gamma Q(s_{t+1},a_{t+1})  \\
% - Q(s_t, a_t) \right)
% \end{split}
% \end{equation}

\begin{equation}
\label{eq:this update}
\begin{aligned}
Q(s_t, a_t) \leftarrow Q(s_t, a_t) + 
\alpha \Bigl( r_{t+1} + \gamma Q(s_{t+1}, a_{t+1}) \\
\qquad\qquad - Q(s_t, a_t) \Bigr)
\end{aligned}
\end{equation}

In Equation~\ref{eq:this update}, $s_t$ is prompt result at step $t$ and $Q(s_t, a_t)$ is the reward value at step $t$, which uses $a_t$ to rewrite prompt section $s_t$. $\alpha$ and $\gamma$ are hyper-parameters.

\section{Experiments}\label{sec:Experiments}

In this section, we demonstrate the effectiveness of our framework through experiments. Firstly we introduce our datasets and training details, then we present the main results of the experiments, and finally show the results of the ablation studies.

\subsection{Datasets and Training Details}\label{sec:Experiment Settings}

To demonstrate the effectiveness of PromptFlow, we conduct our experiments on three real-world datasets: Cluener\citep{xu2020cluener2020} for named entity recognition task, Thucnews\citep{thuctc2016} for classification tasks, and Squad \citep{rajpurkar2018know} for machine reading comprehension task. For each dataset, we randomly select 1,400 data samples as training data, another 600 as test data to validate the performance. 
In our experiment, we run the MSGD optimizer and the MSGD-RL optimizer for each dataset. The model runs 10 iterations by default. We set the beam search size to $6$ for prompt initialization in each section. The value of $k$ is set to 3, meaning that the top 3 prompts are selected for the next iteration. The parameter $\alpha$ is set to $2$, indicating that $2$ simulated annealing instances are used in each iteration to preserve the global optimum. By default, we set the optimization objective to F1 for performing various tasks. Unless otherwise specified, GPT-4 is used as the base model for running experiments.

\begin{table*}[t]
\centering
\caption{Performance across various tasks and prompting methods.}
\footnotesize
\renewcommand{\arraystretch}{1.1}

\begin{tabular}{ccccccccc}
\hline
\multirow{2}{*}& \multirow{2}{*}{\textbf{Model}}
& \multicolumn{5}{c}{\textbf{NER Task}} 
& \multicolumn{1}{c}{\textbf{CLS Task}} 
& \multicolumn{1}{c}{\textbf{MRC Task}} \\
\cline{3-9}
& & Name & Address. & Scene. & Position & \textbf{Avg.} 
& THUCNEWS & SQUAD \\
\hline
& Empty & 61.05 & 46.54 & 51.75 & 67.20 & 55.16 & 70.67 & 70.86 \\
& CoT   & 62.53 & 45.05 & 51.05 & 66.14 & 56.36 & 71.03 & 71.11 \\
\hline
& Manual PE   & 74.67 & 42.55 & 62.02 & 46.42 & 60.53 & 73.06 & 70.34 \\
\hline
& BS  & 62.88 & 48.36 & 61.88 & 70.02 & 57.20 & 71.40 & 70.80 \\
& APE   & 74.04 & 78.10 & 80.02 & 77.96 & 57.50 & 72.00 & \textbf{75.49} \\
& APO   & 73.12 & \textbf{88.94} & 81.94 & 83.20 & 57.71 & 72.20 & 72.92 \\
& OPRO  & 74.66 & 88.72 & 78.55 & 82.11 & 65.05 & 73.33 & 70.89 \\
& PE2   & 73.77 & 87.15 & 83.20 & 82.72 & 63.07 & 72.77 & 71.10 \\
\hline
& \textbf{MSGD}   & 84.32 & 88.17 & 83.20 & \textbf{84.58} & 75.10 & 76.46 & 74.61 \\
& \textbf{MSGD-RL}   & \textbf{84.96} & 88.34 & \textbf{86.61} & 83.33 & \textbf{78.63} & \textbf{81.91} & 75.24 \\
\hline
\end{tabular}
\label{tab:task-performance}
\end{table*}

We compare PromptFlow with the following baselines, including BS\citep{xie2023self}, APE~\citep{zhou2022large}, APO~\citep{pryzant2023automatic}, OPRO~\citep{yang2023large}, and PE2~\citep{ye2023prompt}. The baselines implementation details can be found in Appendix~\ref{sec:Implementation details}.

\subsection{Overall Performance}\label{sec:Experiment Result}
The main results are shown in Table~\ref{tab:task-performance}. Our method consistently outperforms most baselines across multiple tasks, achieving an average improvement of $8.8\%$ in F1-score compared to the strongest baseline OPRO. Notably, our approach demonstrates superior performance in terms of NER task, which achieves a significant gain over OPRO by $13.58\%$. On Classification task, our model outperforms OPRO by $8.58\%$. However, on MRC task, the improvement is relatively modest. This might due to the fact that the main challenge for reading comprehension tasks lies in deep semantic understanding and reasoning, rather than just the input prompts, thus prompt engineering may not significantly enhance LLM’s ability to understand complex contexts. Tasks like entity recognition and classification involve selecting multiple appropriate labels from a predefined set. By designing suitable prompts, LLM can be better guided to focus on key information, such as relationships between labels or label-specific considerations, thereby significantly improving performance. Incorporating RL into MSGD enables our method to better learn and recycle experience in training processes, leading to improved performance compared to MSGD optimizer. Compared to manual prompt engineering, our approach achieves an average improvement of $10.2\%$ in F1-score, primarily due to our gradient-based RL optimization mechanism and a rich library of operators. These results highlight the effectiveness of our PromptFlow in automatic prompt improvement, significantly enhancing the capability of LLMs for solving NLP tasks. For more detailed data results, please refer to Appendix~\ref{sec:More experiment results}.

\begin{figure}[htbp]
  \centering
  \begin{subfigure}[t]{0.49\linewidth} % 左侧图片
   \includegraphics[width=1\linewidth]{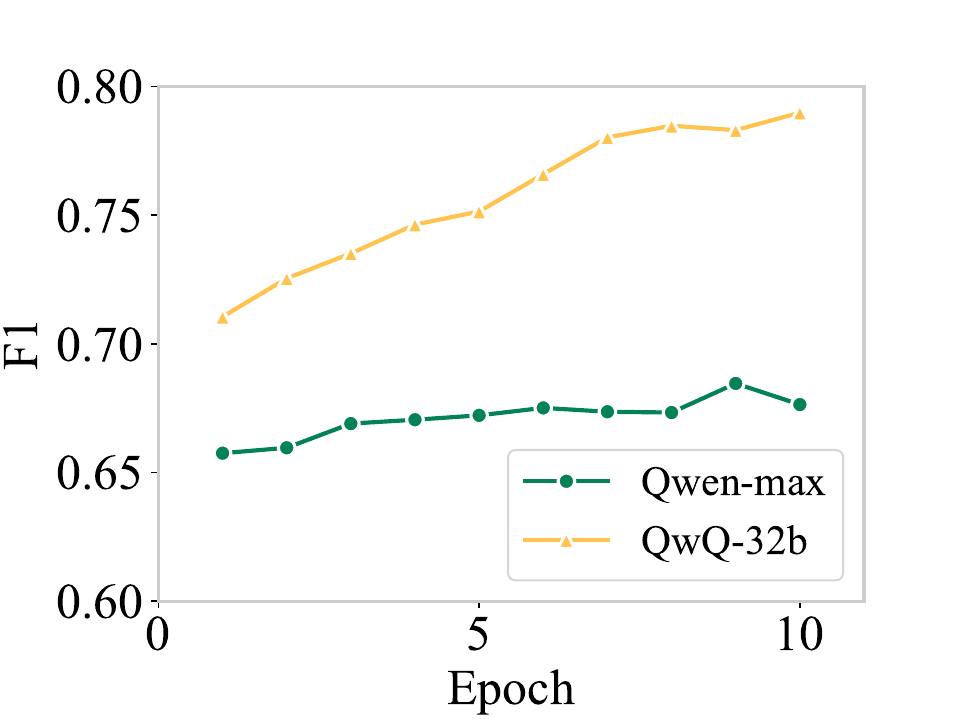} % 替换为你的图片文件名
   \caption{\small{NER Task}}
   \label{fig:Cluener on Qwen}
  \end{subfigure}
  \begin{subfigure}[t]{0.49\linewidth} % 右侧图片
    \includegraphics[width=1\linewidth]{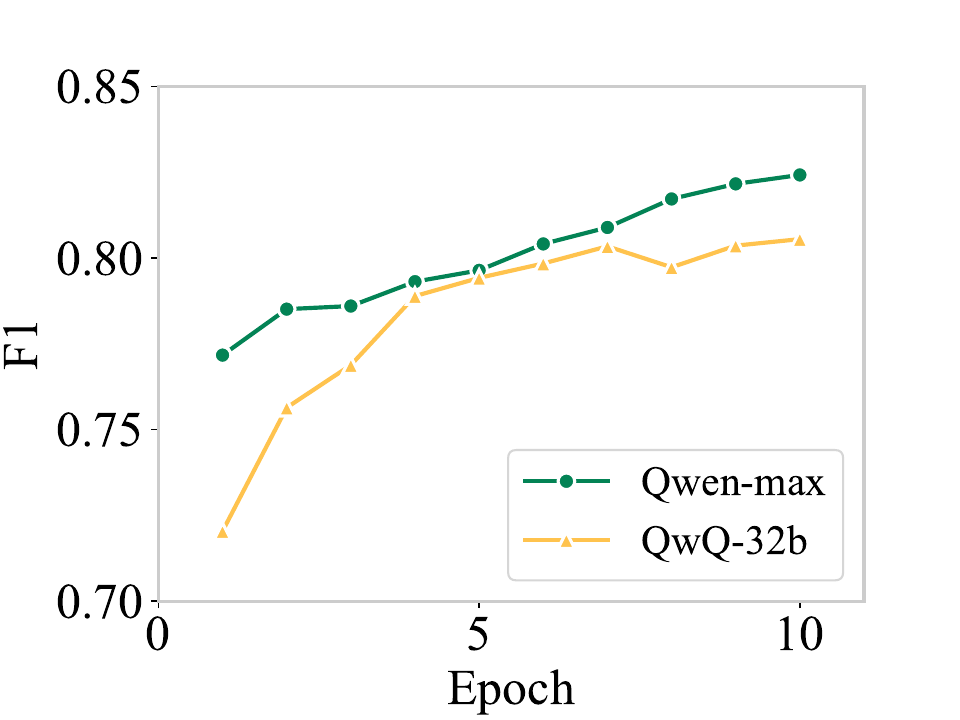} % 替换为你的图片文件名
    \caption{\small{CLS Task}}
    \label{fig:Thucnews on Qwen}
  \end{subfigure}
  \caption{Results on reasoning and non-reasoning models}
  \label{fig:Results on reasoning}
\end{figure}

\subsection{Operator Selection Analysis}
\begin{figure*}[t]
  \centering
  \begin{subfigure}[t]{0.48\linewidth}
    \includegraphics[width=\linewidth]{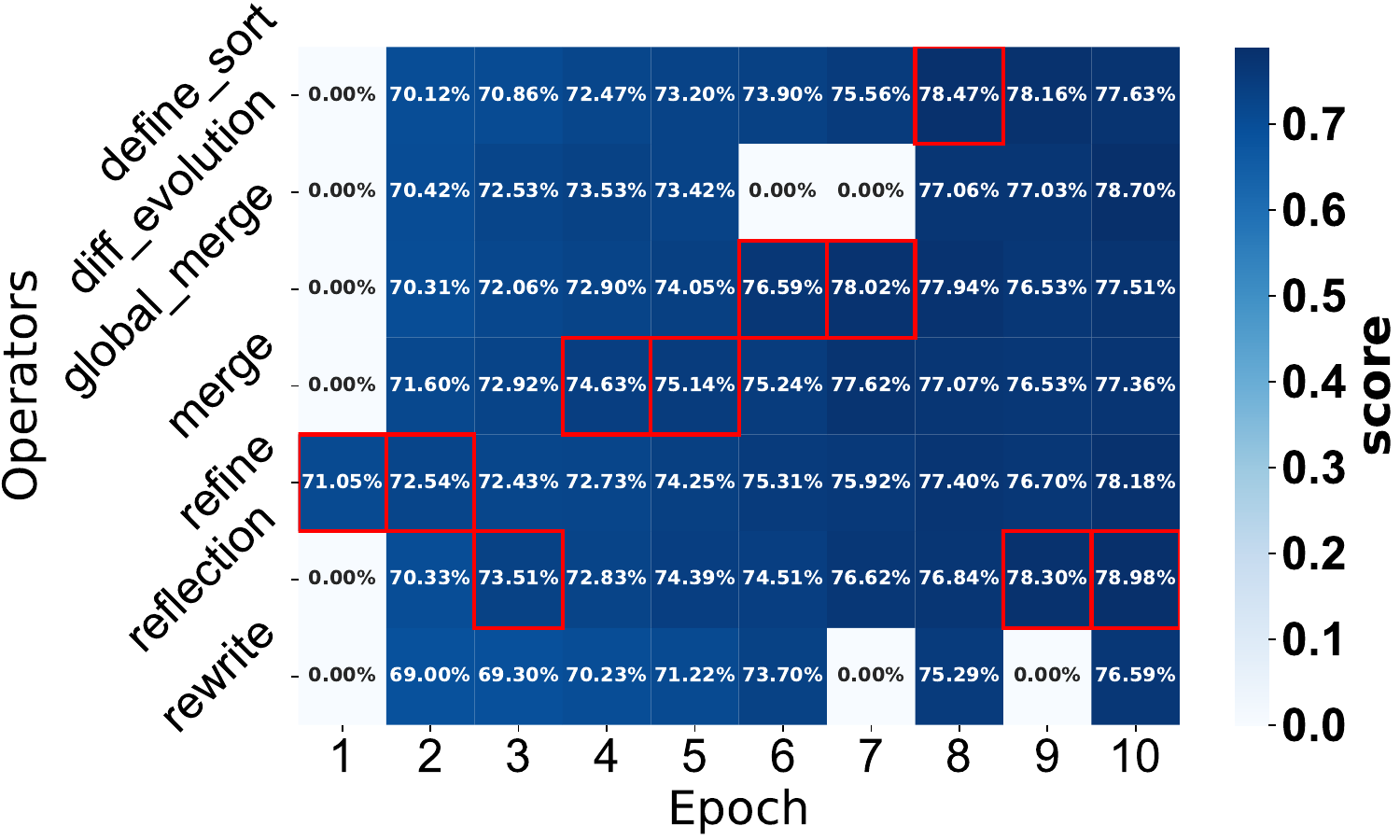}
    \caption{\small{NER Task}}
    \label{fig:Thucnews operator}
  \end{subfigure}
  \begin{subfigure}[t]{0.48\linewidth}
    \includegraphics[width=\linewidth]{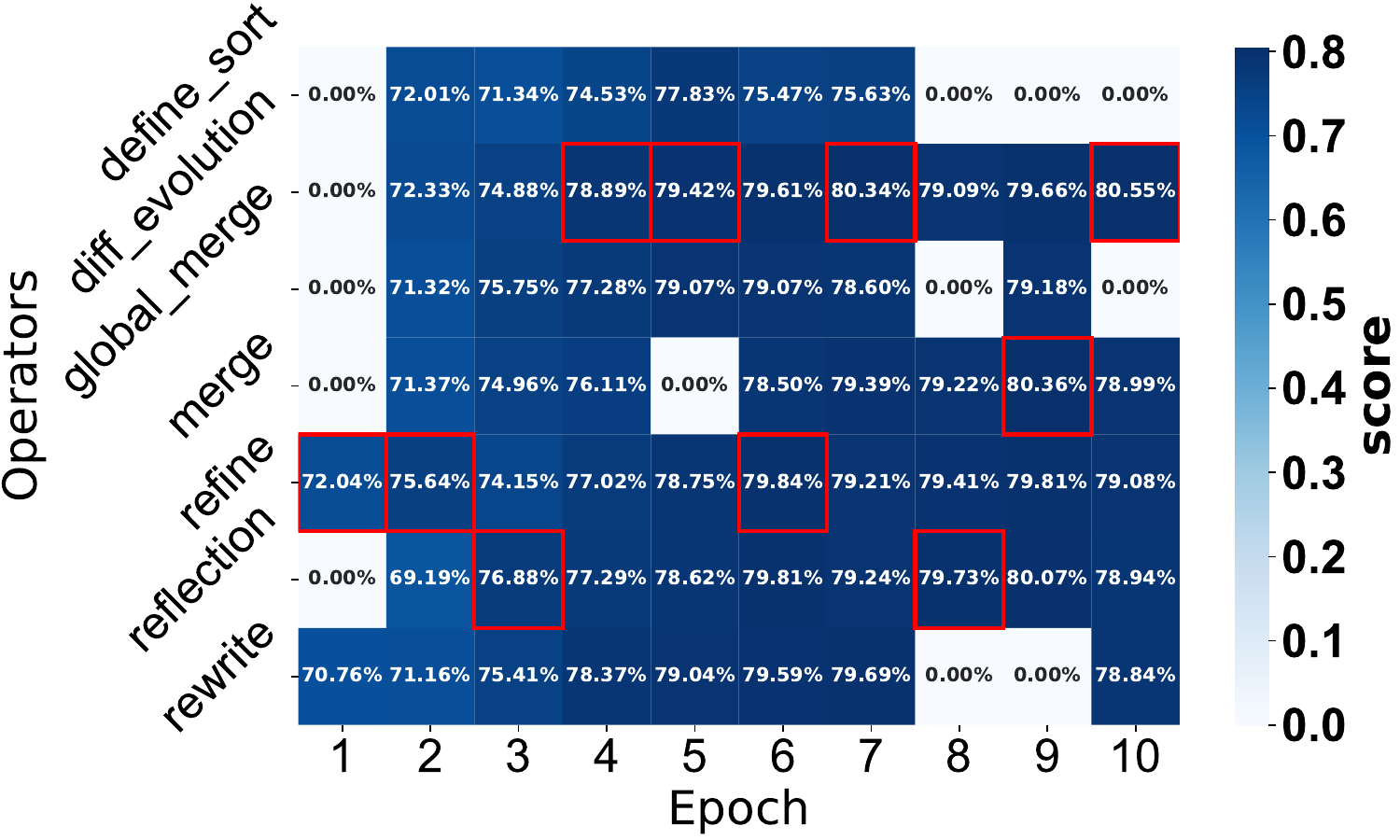}
    \caption{\small{CLS Task}}
    \label{fig:Cluener operator}
  \end{subfigure}
  \caption{Operator selection and performance in each iteration}
  \label{fig:Operators}
\end{figure*}

We present the results of operators selected in each iteration in Figure~\ref{fig:Operators}. In classification task, \textbf{diff evolution} operator achieves relatively better results, while the performance of \textbf{rewrite} is comparatively poor. Classification labels are complex. Local rewriting can alter label meanings and overlook their relationships, limiting improvement. By leveraging global search capabilities, \textbf{differential evolution} demonstrates better results.In NER task,  \textbf{reflection} operator performs the best. Experimental analysis reveals that the NER task is generally more challenging than CLS task. Using the reflection method allows for effective summarization and utilization of difficult cases to guide improvements. Interestingly, \textbf{rewrite} consistently yields the worst performance on both the CLS and NER tasks.This suggests that global prompt rewriting is ineffective, and targeted optimization of meta-level components is more appropriate. At the same time, we also found that task types exhibit distinct preferences for operators. For example, CLS favors diff evolution, while NER tends to prefer reflection. This indicates that different tasks are best optimized using different operators. This experience and knowledge will be recorded during the iteration process. Later, when facing different data for the same task, the system doesn’t need to learn from scratch but can instead initialize based on the tendency probabilities of different operators.

\subsection{Effects of Different Optimization Metrics}
\begin{figure}[t]
  \centering
  \begin{subfigure}[t]{0.41\linewidth}
   \includegraphics[width=1\linewidth]{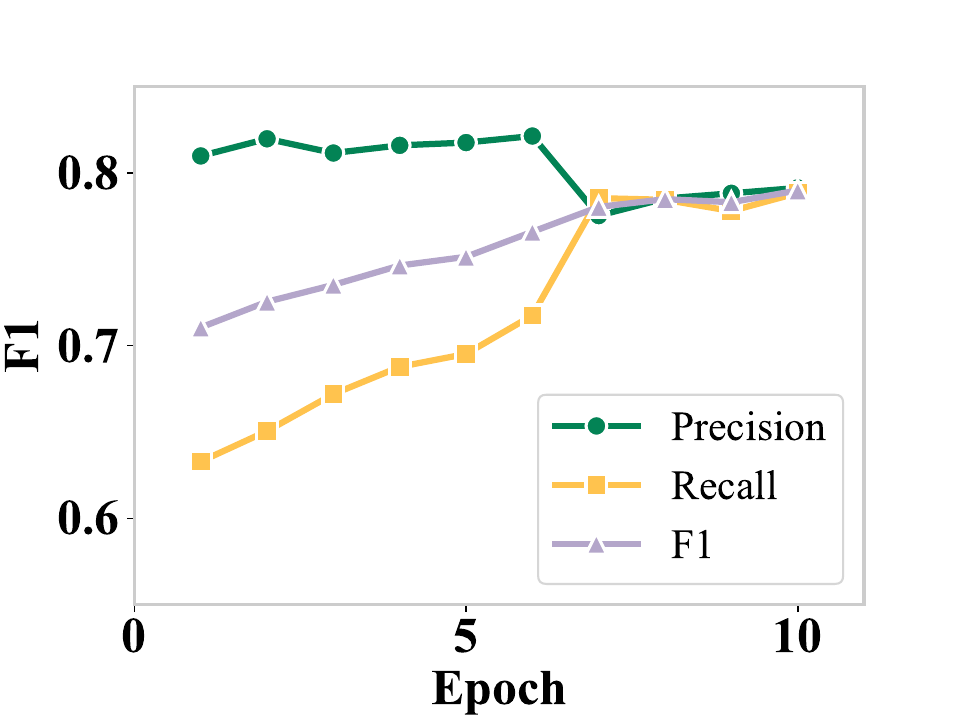}
  \end{subfigure}
  %\hspace{5pt}
  \begin{subfigure}[t]{0.41\linewidth}
    \includegraphics[width=1\linewidth]{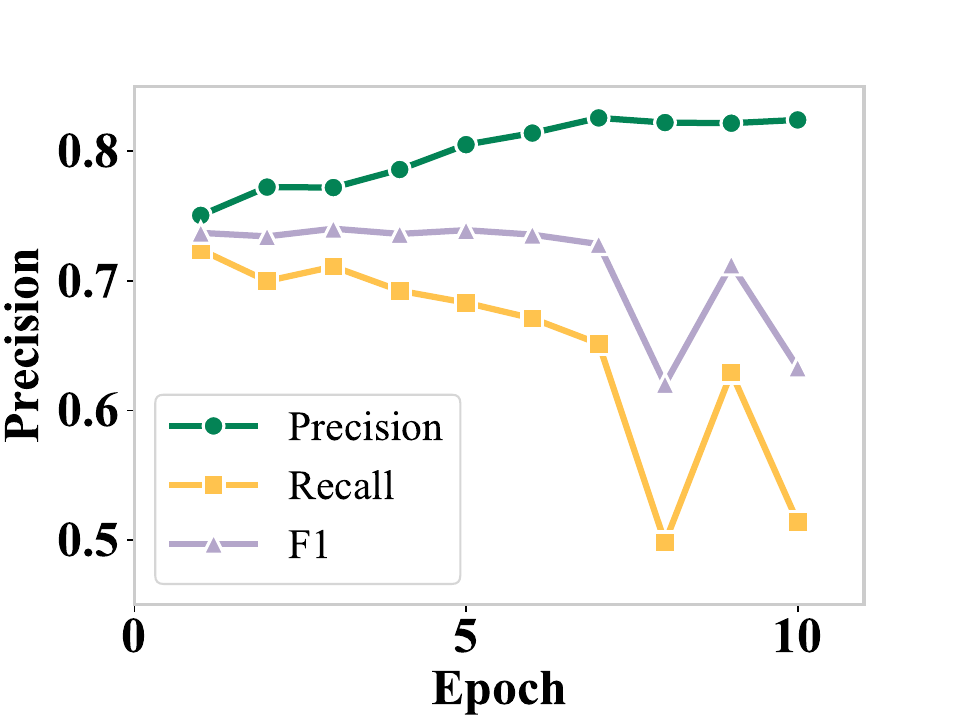}
  \end{subfigure}
  \begin{subfigure}[t]{0.41\linewidth}
    \includegraphics[width=1\linewidth]{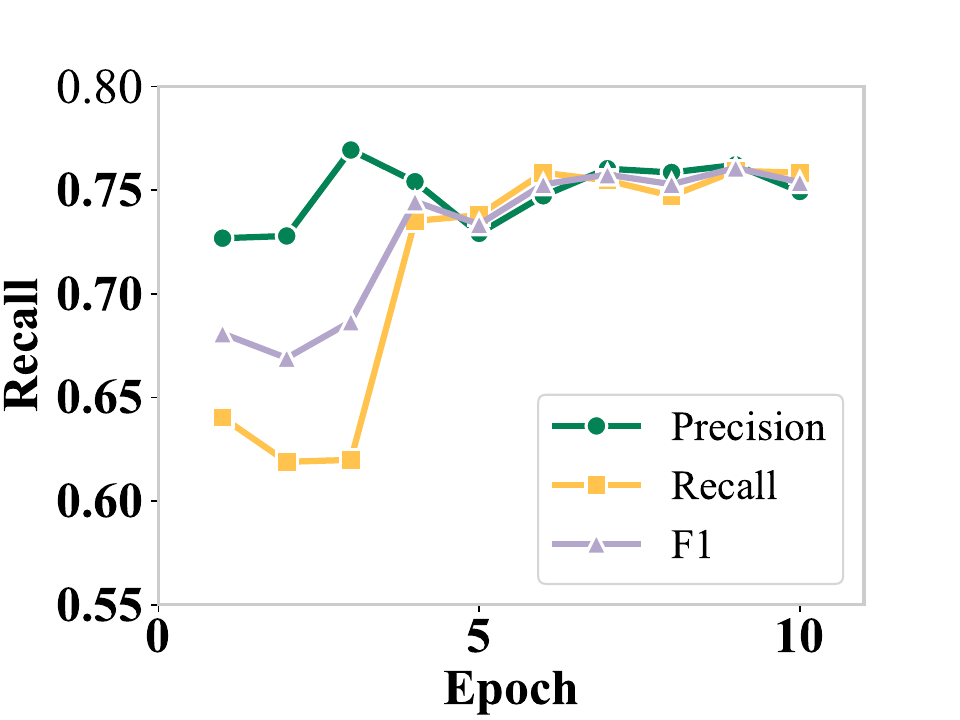}
  \end{subfigure}
  \caption{\small{Analysis of model on different metrics}}
  \label{fig:Analysis metrics}
\end{figure}

Our PromptFlow supports configuring different metrics as optimization targets. In this setup, during the iteration process, the evolution and selection of prompts prioritize retaining results with higher metric values, thereby better aligning with the optimization goal. We compare the model results on CLS task using F1, Precision, and Recall as optimization targets based on GPT-4. The experimental results demonstrate substantial improvements over the initial metrics, thereby validating the effectiveness of our PromptFlow in optimizing for specific evaluation targets. As shown in Figure ~\ref{fig:Analysis metrics}, we have some intriguing discoveries. When optimizing for F1, the model boosts Recall but at the cost of Precision. In contrast, when targeting Precision, both Recall and F1 drop significantly, suggesting that improving Precision is comparatively more challenging. However, when focusing on Recall, we observe a substantial gain in Recall with minimal loss in Precision, which in turn also raises F1. This indicates that F1 and Recall can be good optimization targets, while Precision, to some extent, may reduce F1. Therefore, if your goal is to maintain F1, it would not be a good choice.

\section{Ablation Analysis}
\subsection{Performance on Reasoning Model}
With the emergence of reasoning models, LLM capabilities have seen significant improvements. We compare the performance of promptFlow on both reasoning and non-reasoning models to analyze whether promptFlow still has room for improvement. When we run Promptflow on reasoning(QwQ-32b) and non-reasoning(Qwen-max) models, we have an interesting finding shown in Figure~\ref{fig:Results on reasoning}. As we evaluate the performance on Qwen-Max and QwQ-32b, after a limited number of iterations, QWQ-32b has more significant improvement based on the initial prompt on both the CLS and NER tasks. As a reasoning model, QwQ-32b demonstrates greater sensitivity to prompt adjustments, suggesting that reasoning models may deliver superior performance with PromptFlow. Besides, it can be seen that when using PromptFlow to optimize PE, the reasoning model performs better on more complex tasks (NER), outperforming the non-reasoning model in both final performance and convergence speed.

\subsection{Results on Different Optimizers}\label{sec:Effects of different optimizers}

To demonstrate the effectiveness of optimizers, we compare the results of different optimizers, shown in Figure~\ref{fig:Optimizers performance}. Based on GPT-4, we evaluate the performance using MSGD optimizer and MSGD-RL optimizer respectively. The experimental results demonstrate that the MSGD-RL optimizer outperforms the MSGD optimizer. In our experiment, the MSGD-RL optimizer selects the direction with the steepest loss reduction for optimization, enabling it to approach the optimal solution more efficiently within a limited number of iterations, incorporating reinforcement learning method to learn and recycle experience in training processes. 

\section{Conclusion}\label{sec:Conclusion}
In this paper, we introduce PromptFlow,  a modular training framework for prompt generation. Compared to prior research, PromptFlow enhances prompt performance at the meta-prompt level by enabling optimizers to select from a variety of operators. We involve a gradient-based reinforcement learning optimization mechanism to recycle and leverage experience to enhance the performance. This framework significantly reduces the cost of manually optimizing prompts while offering the flexibility to continuously expand the library of optimizers and operators, seamlessly integrating the latest prompt engineering techniques. Our experimental results demonstrate that PromptFlow achieves outstanding performance across multiple benchmarks, covering a variety of different NLP tasks.We will further discuss the limitations of the paper and future work in the appendix.

\clearpage
\bibliography{aaai2026}

\clearpage
\appendix

\section{Limitations and Future Work}
Our research has several limitations. Firstly, PromptFlow relies on high-quality supervised datasets, making it unsuitable for unsupervised scenarios. Secondly, PromptFlow currently incurs a significantly high computational cost, particularly during the reasoning and evaluation stages, which calls for the integration of more efficient reasoning and evaluation methods. Furthermore, PromptFlow may not have a substantial impact when the initial prompt already achieves a high score with the base LLM or when the task relies more on domain-specific knowledge rather than prompt engineering. \\
Future work can focus on several key aspects. First, efforts should be made to reduce the total cost of each iteration, thereby improving efficiency. Second, the diversity of operators and optimizers can be further enriched to enhance the framework's flexibility and performance. Finally, additional experiments on a wider range of tasks are needed to validate the framework's applicability and determine the optimal configuration choices.

\section{Contextual Background}  \label{sec:Contextual Background}
The design of our framework is inspired by real-world business applications. While current Automatic Prompt Engineering(APE) frameworks are mostly theoretical, our PromptFlow is more of an application-oriented framework. Our design has been rigorously tested in complex, real-world industrial scenarios and has received consistently positive feedback. To further validate its effectiveness, we conduct experiments on open-source datasets.In light of our target application scenario, we place emphasis on evaluating our approach on Chinese-based LLM models and datasets. GPT is additionally employed to ensure a fair comparison with other APE frameworks.While PromptFlow may require a relatively high computational resources at first glance, the cost is reasonable when considering the complexity and practical impact of the target real-world problem. It mitigates the need for labor-intensive manual prompt engineering (PE) optimization commonly encountered in AI deployment. \\
In the future, we plan to conduct further experiments and optimizations on a broader range of open-source English datasets, which may uncover new opportunities for improvement and research.To date, supported by our current experimental results, PromptFlow is proved to be well-suited for Chinese scenarios and exhibits practical applicability.

\section{Implementation Details} \label{sec:Implementation details}
For clarity and ease of reading, all of the Chinese prompts are translated into English here.

\subsection{Implementation of Operators} \label{sec:Details of operator implementation}
These example illustrate how operators are integrated into PromptFlow, demonstrating their roles in modifying meta-prompts.
\begin{tcolorbox}[title=\textbf{PromptFlow Operator: Refine}, colback=white, colframe=black,arc=10pt,breakable]
    You are an excellent prompt engineer, and you are familiar with the "Refine" optimization method, which is to improve or optimize the existing prompts. You can make the prompts better and more accurate through small adjustments or enhancements. \\

    Below is a \texttt{{\{\{Module\}\}}} of a Prompt, the original expression is as follows: \\
    \texttt{{\{\{Module\_Desc\}\}}}\\
    \\
    Please use the "Refine" method to optimize the expression. You need to return the result directly in JSON format, and the JSON value must be in string format: \\

    \texttt{\{} \\
    \texttt{"\texttt{{\{\{Module\}\}}}":""} \\
    \texttt{\}}
\end{tcolorbox}

\begin{tcolorbox}[title=\textbf{PromptFlow Operator Using Case: Refine}, colback=white, colframe=black,arc=10pt,breakable]
\section*{Input}
\textbf{[Module]:} horoscope\\
\textbf{[Module\_Desc]:} Focus on astrology, horoscope analysis, fortune prediction and other horoscope-related content.

\section*{Output}
\textbf{[Module]:} horoscope\\
\textbf{[Module\_Desc]:} Dedicated to the fields of astrology, horoscope character analysis, and personal fortune prediction, encompassing knowledge and culture related to horoscopes.
\end{tcolorbox}

\begin{tcolorbox}[title=\textbf{PromptFlow Operator: Reflection}, colback=white, colframe=black,arc=10pt,breakable]
    \section*{Background}
    You are a natural language processing expert who is good at extracting key information from large amounts of text data and systematically summarizing it. You need to first extract common problems and root causes based on the existing bad case analysis results, and then improve the description of {{module}} in a targeted manner.The analysis results of the bad case you received are as follows:
    
    \texttt{{\{\{Bad\_Case\_Reason\_List\}\}}}

    \section*{Task Description}
    The current description of \texttt{{\{\{Module\}\}}} is as follows:\\
    \texttt{{\{\{Module\_Desc\}\}}}\\
    \\
    Read the bad case analysis results and complete the following tasks for the description of \texttt{{\{\{Module\}\}}}:\\
    1. Common problem extraction: Classify bad cases and extract common problems. You need to highlight high-frequency or high-impact key issues to ensure that the classification dimensions are clear and logically consistent.\\
    2. Root cause analysis: For each type of problem, further analyze its root cause. You need to be brief and concise and get to the point without any vague nonsense.\\
    3. Improvement of \texttt{{\{\{Module\}\}}} description: Optimize the description of \texttt{{\{\{Module\}\}}} based on the above summary. Expand and supplement \texttt{{\{\{Module\}\}}} when necessary, while avoiding redundancy to ensure that the key points are highlighted and the description is accurate.\\
    You need to return the result directly in json format, and the json value must be in string format:\\
    \texttt{\{} \\
    \texttt{"Common problem extraction":"",\\
            "Root cause analysis":"",\\
            "Improved \texttt{{\{\{Module\}\}}} description":""
            } \\
    \texttt{\}}
\end{tcolorbox}

\begin{tcolorbox}[title=\textbf{PromptFlow Operator Use Case: Reflection}, colback=white, colframe=black,arc=10pt,breakable]
\section*{Input}
\textbf{[Module]:} horoscope\\
\textbf{[Module\_Desc]:} Focus on astrology, horoscope analysis, fortune prediction and other horoscope-related content.\\
\textbf{[Bad\_Case\_Reason\_List]:} ["The text primarily addresses how to enhance fortune through furniture arrangement, which falls under Feng Shui and aligns more with home layout optimization than with astrology or Horoscope personality analysis.",\\
"The marking error arises because the text does not address astrology, Horoscope personality analysis, or fortune prediction; instead, it discusses marriage psychology."
...\\
]

\section*{Output}
\textbf{[Common\_Problem\_Extraction]:} "1.Content unrelated to astrology or horoscope analysis is misclassified as "horoscope"; 2.Psychological testing and personality analysis, resembling horoscope definitions, are often miscategorized as "entertainment" or "society"; 3.Topics like home layout and workplace psychology are classified as "horoscope" due to rigid adherence to definitions."\\
\textbf{[Root\_Cause\_Analysis]:} "An inaccurate or overly broad understanding of the definition of horoscope leads to the inclusion of content with no direct connection to astrology within the "horoscope" category; Psychological tests and personality analyses may resemble astrology-related content in form but do not fully align with the core elements of astrology, causing classification confusion; In certain contexts, topics such as home preferences or workplace psychology are forcibly categorized as "horoscope", reflecting a mechanical application of definitions rather than a classification based on the characteristics of the content itself." \\
\textbf{[Improved\_Module\_Description]:} "Focuses on fields related to astrology, horoscope personality analysis, and personal fortune predictions. This includes, but is not limited to, personality trait analysis through horoscope symbols and predictions of future trends based on celestial positions. It also encompasses content that explores personal traits and behavioral patterns using similar methods (such as psychological testing), but excludes purely mental health advice, home layout guidance, or workplace skills sharing."\\
\end{tcolorbox}

\subsection{Details of Meta-Prompt Initialization}  \label{sec:Details of prompt initialization}
Here we show the initial prompts used in Classification, NER, and Task respectively, which help readers understand how the prompts are composed of meta-prompts. These examples illustrate that PromptFlow is capable of performing prompt optimization in two modes: (1) starting from scratch with minimal human input, (2) starting from an initial template with provided initial definitions. This dual capability ensures broad adaptability across different practical scenarios.

\begin{tcolorbox}[title=\textbf{Initial Prompt: from template}, colback=white, colframe=black,arc=10pt,breakable]

\section*{Task Description}
You are a text classification model specialized in recognizing and categorizing Chinese news content. Your task is to assign the given news text to the appropriate news category label.

\section*{Label Descriptions}

\begin{enumerate}[label=(\arabic*)]
    \item \textbf{Games} Definition: XXX \\
    \textit{}
    
    \item \textbf{Real} Definition: XXX \\
    \textit{}
    
    \item \textbf{Education} Definition: XXX \\
    \textit{}

    \item \textbf{Sports} Definition: XXX \\
    \textit{}

    \item \textbf{Horoscope} Definition: XXX \\
    \textit{}

    \item \textbf{Home} Definition: XXX \\
    \textit{}
    
    \item \textbf{Lottery} Definition: XXX \\
    \textit{}

    \item \textbf{Entertainment} Definition: XXX \\
    \textit{}

    \item \textbf{Current Affairs} Definition: XXX \\
    \textit{}

    \item \textbf{Technology} Definition: XXX \\
    \textit{}

    \item \textbf{Society} Definition: XXX \\
    \textit{}
    
\end{enumerate}

\section*{Examples}
\texttt{XXX}

\section*{Step-by-step}
\texttt{XXX}

\section*{Input}
\texttt{{\{\{Input\}\}}}

\section*{Output}
\texttt{{\{"label":""\}}} \\

\end{tcolorbox}

\begin{tcolorbox}[title=\textbf{Initial Prompt: from scratch}, colback=white, colframe=black,arc=10pt,breakable]

\section*{Task Description}
You are a natural language processing expert. You need to handle a task of extracting key information from user text, which involves the classification of multiple labels. This task aims to extract entity information.

\section*{Entity Description}
\begin{enumerate}[label=(\arabic*)]
    \item \textbf{Address} Definition: XXX (model will decide waht section) \\
    \textit{}
    
    \item \textbf{YYY} Definition: XXX \\
    \textit{}
    
    ... (model will decide add what section) \\
    \textit{}
    
    \item \textbf{ZZZ} Definition: XXX \\
    \textit{}
    
\end{enumerate}

\section*{Few Shot}

\section*{...}

\section*{Example}

\section*{Input}
\texttt{{\{\{Input\}\}}}

\section*{Output}
return the results directly in JSON format. \\
\texttt{\{} \\
\texttt{"x1":\{"xxx": [[x, x]]\},} \\
\texttt{"...":\{"xxx": [[x, x]]\},} \\
\texttt{"xn":\{"xxx": [[x, x]]\},} \\
\texttt{\}} \\

\end{tcolorbox}

\subsection{Case Study}
\subsubsection{MSGD Optimizer} 
We provide a detailed explanation of how the MSGD Optimizer works. Taking the NER task as a case study, based on batch dataset $D=\{(x_0, y_0),(x_1, y_1), ..., (x_m, y_m)\}$ , we first evaluate the performance of the initial prompt to establish our starting point.

\begin{tcolorbox}[title=\textbf{Initial Evaluation}, colback=white, colframe=black, arc=10pt, breakable]

\section*{Meta-prompt Module}
\begin{center}
\begin{tabular}{|l|c|c|}
\hline
\textbf{Initial Prompt} & \textbf{Score} & \textbf{Loss} \\
\hline
\textbf{P} & 0.64513 & 0.35487\\
\hline
\end{tabular}
\end{center}

\end{tcolorbox}

Next,we identify the set of available Operators $O = \{o_0, o_1, ..., o_n\}$ and the meta-prompt $S = \{s_0, s_1, ..., s_l\}$. Each prompt operator and the meta-prompt module are assigned an initial score, which reflects their initial selection probability during the iteration process. This scoring mechanism allows us to quantify the effectiveness and guide the optimization procedure.An illustrative example is shown below,
\begin{tcolorbox}[title=\textbf{Transition Matrix}, colback=white, colframe=black, arc=10pt, breakable]

\begin{center}
\begin{tabular}{|c|c|c|c|c|}
\hline
\textbf{Meta-P} & \textbf{Rewrite} & \textbf{Refine} & \textbf{Reflect} & \textbf{COT} \\
\hline
\textbf{Address} & 0.0625 & 0.0625 & 0.0625 & 0.0625 \\
\hline
\textbf{Book}    & 0.0625 & 0.0625 & 0.0625 & 0.0625 \\
\hline
\textbf{Name}    & 0.0625 & 0.0625 & 0.0625 & 0.0625 \\
\hline
\textbf{Company} & 0.0625 & 0.0625 & 0.0625 & 0.0625 \\
\hline
\end{tabular}
\end{center}

\end{tcolorbox}

At the beginning of training, we randomly select a subset of Operators and meta-prompt module for optimization. The selected operators and meta-prompts for the current iteration are as follows. This means we utlize \textbf{rewrite} operator to update label \textbf{address} section and \textbf{refine} to optimize label \textbf{book}.

\begin{tcolorbox}[title=\textbf{Chosen Operators and Meta-Prompts}, colback=white, colframe=black, arc=10pt, breakable]
\begin{center}
\begin{tabular}{|l|}
\hline
(Address,    Rewrite) \\
\hline
(Book,       Refine) \\
\hline
\end{tabular}
\end{center}
\end{tcolorbox}

Next, we employ our Evaluator to assess the performance of the updated prompts on the dataset. Using real-world data, we compute the discrepancy between the model's predictions and the ground truth labels, which serves as our loss function $L$ for optimization. Based on prior performance, we can calculate the uplift or downlift as the $G$. If the loss increases at this step, the selection probability of the corresponding operator should be decreased; conversely, if the loss decreases, the selection probability should be increased accordingly.

\begin{tcolorbox}[title=\textbf{Calculate Score}, colback=white, colframe=black, arc=10pt, breakable]
\centering
\footnotesize
\setlength{\tabcolsep}{8pt}
\begin{tabular}{|l|l|c|}
\hline
\textbf{Pairs} & \textbf{Field} & \textbf{Value} \\
\hline
\multirow{4}{*}{(Address, Rewrite)} & Init Score    & 0.64513 \\
                                      & Init Loss    & 0.35487 \\
                                      & Cur Score     & 0.66812 \\
                                      & Cur Loss    & 0.33188 \\
                                      & Gradient      & +0.02299 \\
\hline
\multirow{4}{*}{(Book, Refine)}     & Init Score    & 0.64513 \\
                                      & Init Loss    & 0.35487 \\
                                      & Cur Score     & 0.62440 \\
                                      & Cur Loss     & 0.37560 \\
                                      & Gradient      & -0.02073 \\
\hline
\end{tabular}
\end{tcolorbox}

Then, we normalize the loss using $Norm$. Assuming $\alpha$ = 1, we can now calculate the value of $Q(s_i, o_j)=Q(s_{i-1}, o_{j-1})*(1+\alpha Norm)$

\begin{tcolorbox}[title=\textbf{Prob Calculation}, colback=white, colframe=black, arc=10pt, breakable]
\begin{center}
\small
\begin{tabular}{|l|c|c|}
\hline
\textbf{Pairs} & \textbf{Norm} & \textbf{$Q(s_i, o_j)$}\\
\hline
(Address, Rewrite) & 0.03564 & 0.0647  \\
\hline
(Book, Refine) & -0.03213 & 0.0605\\
\hline
\end{tabular}
\end{center}

\end{tcolorbox}

Therefore, we update the probability of selecting each operator and meta-prompt module.

\begin{tcolorbox}[title=\textbf{Update Transition Matrix}, colback=white, colframe=black, arc=10pt, breakable]

\begin{center}
\begin{tabular}{|c|c|c|c|c|}
\hline
\textbf{Meta-P} & \textbf{Rewrite} & \textbf{Refine} & \textbf{Reflect} & \textbf{COT} \\
\hline
\textbf{Address} & 0.0647 & 0.0625 & 0.0625 & 0.0625 \\
\hline
\textbf{Book}    & 0.0625 & 0.0605 & 0.0625 & 0.0625 \\
\hline
\textbf{Name}    & 0.0625 & 0.0625 & 0.0625 & 0.0625 \\
\hline
\textbf{Company} & 0.0625 & 0.0625 & 0.0625 & 0.0625 \\
\hline
\end{tabular}
\end{center}

\end{tcolorbox}

In the subsequent iteration,based on performance across the past three iterations, we select the best-performing operators and meta-prompt modules to edit new prompt $p^{*}$ . The ultimate objective is to minimize the overall loss, thereby maximizing the model's overall performance metric.

\subsubsection{MSGD-RL Optimizer} 
As seen above, we need to initialize the selection probabilities for both the operators and the meta-prompt sections. To learn and recycle experience from past \textbf{transition matrix}, we define the procession of using operator to refine the section of prompt as a Markov Decision Process (MDP). The state $S$ is the current prompt. The action $A$ is to choose operator $a_j$ to rewrite prompt section $s_i$. The reward $R$ is determined by evaluating the predicted result against the ground truth. The $S^{\prime}$ is the new prompt after rewriting. The $A^{\prime}$ is next action to refine the prompt. Specially, a function $Q_{i,j}^{t}$ is designed to evaluate the value of using $o_j$ to optimize $p_i$ at the current time $t$, thereby supporting decision-making on the optimal action to take, as shown in Equation~\ref{eq:q_max_matrix},where $Q_(i, j)=Q_{L}(s_i, a_j)$ denotes the $Q$-value for state $s_i$ and action $a_j$ at Level $Q$.

\begin{equation}
\label{eq:q_max_matrix}
Q_{i,j}^{t} =
\begin{bmatrix}
Q_(0, 0) & Q(0, 1) & \cdots & Q(0, j) \\
Q(1, 0) & Q(1, 1) & \cdots & Q(1, j) \\
\vdots & \vdots & \ddots & \vdots \\
Q(i, 0) & Q(i, 1) & \cdots & Q(i, j)
\end{bmatrix}
\end{equation}

\begin{tcolorbox}[title=\textbf{Init From Prev Transition Matrix}, colback=white, colframe=black, arc=10pt, breakable]

\begin{center}
\begin{tabular}{|c|c|c|c|c|}
\hline
\textbf{Meta-P} & \textbf{Sort} & \textbf{Refine} & \textbf{Reflect} & \textbf{COT} \\
\hline
\textbf{Address} & 0.0647 & 0.0625 & 0.0625 & 0.0625 \\
\hline
\textbf{Book}    & 0.0605 & 0.0605 & 0.0820 & 0.0625 \\
\hline
\textbf{Name}    & 0.0610 & 0.0625 & 0.0625 & 0.0440 \\
\hline
\textbf{Company} & 0.0625 & 0.0550 & 0.0625 & 0.0625 \\
\hline
\end{tabular}
\end{center}

\end{tcolorbox}

Simultaneously, we use the SARSA~\cite{rummery1994line} algorithm to update the Q-function, recycling experience as well as updating action policy with online learning. Based on the track $\left ( S, A, R, S^{\prime }, A^{\prime}   \right)$, we use $Q(s_{t+1}, a_{t+1})$ to update $Q(s_t, a_t)$.

\begin{equation}
\label{eq:update}
\begin{split}
Q(s_t, a_t) \leftarrow Q(s_t, a_t) + \\
\alpha \left( R_{t+1} + \gamma Q(s_{t+1}, a_{t+1}) - Q(s_t, a_t) \right)
\end{split}
\end{equation}

\begin{tcolorbox}[title=\textbf{Sample Next Actions}, colback=white, colframe=black, arc=10pt, breakable]
\begin{center}
\begin{tabular}{|l|}
\hline
(Name,    Reflect) \\
\hline
(Book,       Sort) \\
\hline
(Company,       Refine) \\
\hline
(Book,       COT) \\
\hline
(Name,       COT) \\
\hline
\end{tabular}
\end{center}
\end{tcolorbox}

In Equation~\ref{eq:update}, $s_t$ is prompt result at step $t$ and $Q(s_t, a_t)$ is the reward value at step $t$, which uses $a_t$ to rewrite prompt section $s_t$. $\alpha$ and $\gamma$ are hyper-parameters, which are defined as 0.5 here. The reward $R_{t+1}$ is obtained by calculating the average of the actions in the above sample.

\begin{tcolorbox}[title=\textbf{Calculate Score}, colback=white, colframe=black, arc=10pt, breakable]
\begin{center}
\small
\begin{tabular}{|l|c|c|}
\hline
\textbf{Pairs} & \textbf{Prev Score} & \textbf{Prev Loss} \\
\hline
(Name, Reflect) & 0.6501 & 0.3499\\
\hline
(Book, Sort) & 0.6052 & 0.3975\\
\hline
(Company, Refine) & 0.5565 & 0.4435\\
\hline
(Book, COT) & 0.5254 & 0.4746\\
\hline
(Name, COT) & 0.6400 & 0.3600\\
\hline

\end{tabular}
\end{center}

\begin{center}
\small
\begin{tabular}{|l|c|c|c|}
\hline
\textbf{Pairs} & \textbf{Cur Score} & \textbf{Loss} & \textbf{Gradient} \\
\hline
(Name, Reflect) & 0.6602 & 0.3398 & 0.0101\\
\hline
(Book, Sort) & 0.6040 & 0.3948 & -0.0012\\
\hline
(Company, Refine) & 0.6512 & 0.4435 & 0.0947\\
\hline
(Book, COT) & 0.5111 & 0.4889 & -0.0143\\
\hline
(Name, COT) & 0.6420 & 0.3580 & -0.002\\
\hline
\end{tabular}
\end{center}
\end{tcolorbox}

Using MSGD-RL Optimizer, the entire PromptFlow can be viewed as an agent. 
Just as in standard RL setups, we need to determine which operators and modules used for optimization in order to maximize the overall objective function.

\begin{tcolorbox}[title=\textbf{Prob Calculation}, colback=white, colframe=black, arc=10pt, breakable]
\begin{center}
\small
\begin{tabular}{|l|c|c|c|}
\hline
\textbf{Pairs} & \textbf{$Q(s_{i+1}, o_{j+1})$} & $R_{t+1}$ & \textbf{$Q(s_i, o_j)$} \\
\hline
(Name, Reflect) & 0.0726 & 0.0174 & 0.0581 \\
\hline
(Book, Sort) & 0.0593 & 0.0174 & 0.0537 \\
\hline
(Comp, Refine) & 0.1497 & 0.0174 & 0.0736 \\
\hline
(Book, COT) & 0.0482 & 0.0174 & 0.0520 \\
\hline
(Name, COT) & 0.0419 & 0.0174 & 0.0412 \\
\hline
\end{tabular}
\end{center}

\end{tcolorbox}

\subsection{Details for Implementation of Baselines}
\label{sec:Details for implementation of baselines}

\paragraph{APE} Inspired by classical program synthesis and human-driven prompt engineering, ~\citet{zhou2022large} introduces Automatic Prompt Engineering (APE) for the automated generation and selection of instructions. This method treats the instruction as a "program" to be optimized, searching through a pool of instruction candidates generated by an LLM to maximize a selected scoring function.

\paragraph{APO} ~\citet{pryzant2023automatic} proposes Prompt Optimization with Textual Gradients (ProTeGi). This method mimics the steps of gradient descent to perform non-parametric "gradient descent" in text-based conversations. Finally, the best prompt is selected through efficient beam search and the best arm identification algorithm.

\paragraph{OPRO} OPRO~\citep{yang2023large} does not directly modify the prompt based on textual feedback or require the prompt to adhere to the same semantic constraints compared to APE and APO. Instead, it optimizes on the basis of the previous generated prompts and their scores (i.e., the optimization trajectory), allowing it to identify commonalities among high-scoring prompts.

\paragraph{PE2} ~\citet{ye2023prompt} introduces a trainable attention gate that balances standard cross-attention with attention over retrieved keys from a datastore within a single layer. However, as the public implementation\footnote{\url{https://github.com/google-research/meliad}} of this method is "not officially supported" and lacks full reproducibility, we approximated it by using attention over the datastore.

\subsubsection{Implementation of Baselines} \label{sec:baseline implementation}

Since we are comparing to several baselines, we will not delve into all the details here. Instead, we provide a brief overview of the prompts used across the different baselines.
\begin{tcolorbox}[title=\textbf{APE}, colback=white, colframe=black,arc=10pt,breakable]

\section*{Prompt}
Your task is to assign the given news text to the appropriate news category label: Fashion, Home, Education, Stocks, Entertainment, Lottery, Society, Real Estate, Horoscope, Technology, Finance, Current Affairs, Games, Sports.

\section*{Text}
\texttt{{\{\{Text\}\}}}

\section*{Output}
\texttt{{\{"label":""\}}} \\

\end{tcolorbox}

\begin{tcolorbox}[title=\textbf{APO}, colback=white, colframe=black,arc=10pt,breakable]

\section*{Task}
Classify the news text into one of these categories: Fashion, Home, Education, Stocks, Entertainment, Lottery, Society, Real Estate, Horoscope, Technology, Finance, Current Affairs, Games, or Sports.

\section*{Text}
\texttt{{\{\{Text\}\}}}

\section*{Examples}
\texttt{text:"A well-known gaming company has announced the launch of a new multiplayer online role-playing game, Realm of Legends, set to release next month.", "label":"Games"}

\section*{Output}
\texttt{{\{"label":""\}}} \\

\end{tcolorbox}

\begin{tcolorbox}[title=\textbf{OPRO}, colback=white, colframe=black,arc=10pt,breakable]

\section*{Task}
Begin by carefully analyzing the news text to identify key themes and topics. Consider the defining characteristics of each news category—such as Fashion, Technology, or Sports—and determine which category best matches the content. Apply classification techniques or criteria to assign the text to the most relevant label. Finally, verify that the chosen category accurately reflects the news text to ensure precise labeling.
categories: [Fashion, Home, Education, Stocks, Entertainment, Lottery, Society, Real Estate, Horoscope, Technology, Finance, Current Affairs, Games, Sports]

\section*{Text}
\texttt{{\{\{Text\}\}}}

\section*{Output}
\texttt{{\{"label":""\}}} \\

\end{tcolorbox}

\begin{tcolorbox}[title=\textbf{PE2}, colback=white, colframe=black,arc=10pt,breakable]

\section*{Task}
Let’s approach this task by carefully examining the news text in detail. Focus on the main topics and themes presented, then methodically determine which category—such as Fashion, Technology, or Sports—best fits the content. Step by step, assign the appropriate label to ensure accurate classification. labels : Fashion, Home, Education, Stocks, Entertainment, Lottery, Society, Real Estate, Horoscope, Technology, Finance, Current Affairs, Games, Sports

\section*{Text}
\texttt{{\{\{Text\}\}}}

\section*{Output}
\texttt{{\{"label":""\}}} \\

\end{tcolorbox}

\section{More Experiment Results}\label{sec:More experiment results}

\subsection{Experiment Results for Different Tasks}
In this paper, as shown in Table~\ref{tab:task-performance}, we present the results of GPT-4 for baseline comparison. We also run PromptFlow on Qwen-max and QWQ-32b, the Chinese-based LLM to compare performance on \textbf{non-reasoning model} and \textbf{reasoning model}. Here we provide all the detailed results, including both train and test accuracies, all expressed in F1 scores.
%每个llm一个表，展示 列为train epoch, test acc. 行为label标签的数据
\subsubsection{NER Task}

Table~\ref{tab:Cluener on Trainset&Testset} presents the results for the training and test sets of QwQ-32b, respectively. Overall, we achieve a $7.93\%$ uplift on the training set, with the label "address" showing the highest improvement at $58\%$. Boldface in the table shows that PromptFlow iteratively uplifts the label score contributing most to the final result, enhancing overall performance. On the test set, the overall score improved by $4.1\%$. By observing the performance of each label on the test set and comparing it with their performance on the training set, we can see that the labels which show improvement on the training set also demonstrate corresponding gains on the test set. Therefore, it can be concluded that the training results align with expectations. The overall score differences here may due to an imbalance in the distribution of samples between the training and test sets.

\begin{table*}[htbp]
\centering
\caption{Cluener Performance on Trainset and Testset}
\label{tab:Cluener on Trainset&Testset}
\small % 调整字体大小以适应表格
\setlength{\tabcolsep}{3pt}
\begin{tabular}{l l *{10}{c}} % 左对齐第一列，其余列居中对齐
    \toprule
    \multirow{2}{*}{Model} &\multirow{2}{*}{Category} & \multicolumn{5}{c}{Trainset} & \multicolumn{5}{c}{Testset}\\
    \cmidrule(lr){3-7}\cmidrule(lr){8-12}
    & & 1     & 3      & 5      & 7     & 10   & 1     & 3      & 5      & 7     & 10   \\
    \midrule
    \multirow{10}{*}{Qwq-32b}&address & 6.25\%   & 28.77\% & 20.44\%  & \textbf{64.88}\%  & 64.93\% & 18.27\%  & 34.06\%  & 33.04\%  & 46.90\%  & 37.40\% \\
&book    & 62.92\%  & 62.92\% & 65.93\% & \textbf{76.92}\%  & 82.22\% & 46.15\%  & 48.72\%  & 47.22\%  & 53.49\%  & 57.47\% \\
&company & 78.79\% & 78.91\% & 80.78\%  & 77.89\%  & 81.08\% & 68.39\%  & 67.40\%  & 63.96\% & 71.35\%  & 74.17\% \\
&game    & 85.97\%  & 87.66\% & 86.92\%  & 86.44\%  & 87.87\% & 78.07\%  & 78.14\% & 75.69\% & 80.58\%  & 80.29\% \\
&government  & 68.87\%  & 70.00\%  & 73.75\%  & 73.29\%  & 72.96\% & 59.49\% & 63.05\%  & 60.40\%  & 60.00\%  & 56.72\% \\
&movie  & 82.76\%  & 78.65\%  & 83.52\%  & \textbf{88.42}\%  & 86.60\%  & 85.72\%  & 85.50\%  & 83.97\%  & 86.57\%  & 85.51\% \\
&name   & 85.39\%  & 87.47\%  & 87.87\%  & 88.00\%  & 88.00\%   & 83.50\%  & 81.89\%  & 81.54\%  & 83.00\%  & 86.70\% \\
&organization  & 73.19\%  & 67.88\%  & 72.92\%  & \textbf{79.59}\%  & 80.53\% & 53.60\%  & 47.48\%  & 28.47\%  & 67.39\%  & 62.37\% \\
&position     & 59.91\%  & \textbf{73.28}\%  & 73.11\% & 71.49\%  & 68.09\%   & 55.32\%  & 59.26\%  & 55.70\% & 55.75\%  & 53.53\% \\
&scene   & 63.64\%  & 65.17\%  & 70.73\%  & 67.53\%  & \textbf{75.00}\% & 59.30\%  & 62.58\%  & 62.43\%  & 56.96\%  & 63.10\% \\
&overall & 71.05\%  & 73.51\%  & 75.14\%  & 78.02\%  & 78.98\% & 62.91\%  & 63.90\%  & 60.49\%  & 67.51\%  & 67.01\%\\
\midrule
    \multirow{10}{*}{Qwen-max}&address & 29.78\%  & \textbf{50.44}\%  & 47.81\%  & 48.35\%  & 48.67\% & 37.66\%  & 52.81\%  & 51.53\%  & 52.17\%  & 51.35\%\\
&book    & 45.92\%  & 46.23\%  & 50.00\%  & 46.15\%  & 44.90\% & 50.67\%  & 49.35\%  & 45.33\%  & 45.07\%  & 51.35\%\\
&company & 73.15\%  & 69.31\%  & 70.86\%  & 73.14\%  & 75.21\% & 71.47\%  & 70.03\%  & 70.12\%  & 70.93\%  & 70.76\%\\
&game    & 78.96\% & 78.91\%  & 79.38\%  & 78.77\%  & 78.88\% & 77.21\%  & 78.65\%  & 78.07\%  & 76.01\%  & 82.71\%\\
&government   & 58.15\% & 59.54\%  & 59.18\%  & 62.80\% & 58.77\% & 63.68\%  & 60.61\%  & 63.59\% & 63.96\%  & 63.92\%\\
&movie   & 84.68\%  & 85.14\%  & 84.30\%  & 83.13\%  & 84.55\% & 86.36\%  & 85.50\%  & 84.85\%  & 87.69\%  & 87.02\%\\
&name    & 84.35\% & 84.66\%  & 85.55\%  & 84.65\%  & 84.77\% & 85.15\% & 85.21\%  & 85.00\%  & 85.43\%  & 84.21\% \\
&organization & 55.64\% & 56.98\%  & 57.97\%  & 60.76\%  & 61.99\% & 62.73\%  & 60.22\% & 62.27\% & 63.44\%  & 63.22\%\\
&position     & 61.71\%  & 59.97\%  & 58.65\%  & 55.35\%  & 56.88\%  & 60.28\%  & 59.03\%  & 58.09\%  & 55.15\%  & 55.15\%\\
&scene   & 62.30\%  & 63.74\%  & 63.84\%  & 62.22\%  & 59.49\% & 58.82\%  & 54.55\%  & 54.09\%  & 58.60\%  & 55.35\%\\
&overall & 65.76\%  & 66.90\% & 67.22\%  & 67.36\%  & 67.64\% & 67.12\% & 67.04\%  & 67.20\%  & 67.59\%  & 67.84\%\\
    \bottomrule
\end{tabular}
\end{table*}

Table~\ref{tab:Cluener on Trainset&Testset} also presents the results for the training and test sets of Qwen-max, respectively. The overall performance on Qwen-Max is inferior to that of QwQ-32b. These results are consistent with our conclusions in the paper, where we note that PromptFlow is more sensitive when applied to a reasoning model. On Qwen-Max, it is difficult to achieve significant improvements on Cluener, as the overall uplift on the training set is $1.88\%$. This is likely due to the inherent complexity of the NER task. Further optimization can be explored in future work.

\subsubsection{Classification Task}
%每个llm一个表，展示 列为train epoch, test acc. 行为label标签的数据
%再解释一下
Table~\ref{tab:Thucnews Trainset and Testset} presents the results for the training and test sets of QwQ-32b, respectively. Overall, we achieve a $8.51\%$ uplift on the training set, with the label "Sports" showing the highest improvement at $27.7\%$. The label "Sports" shows rapid improvement in the early stages of training, but the gains level off as training progressed. Although the initial score for the label "Society" is quite low, its improvement is limited, likely because its definition is relatively abstract and requires stronger contextual comprehension. On the test set, the overall score improved by $6.63\%$. The improvements of the labels are consistent with those observed on the training set.

\begin{table*}[h]
\centering
\caption{Thucnews Performance on Trainset and Testset}
\label{tab:Thucnews Trainset and Testset}
\setlength{\tabcolsep}{3pt}
\begin{tabular}{l l *{10}{c}} % 左对齐第一列，其余列居中对齐
    \toprule
    \multirow{2}{*}{Model} &\multirow{2}{*}{Category} & \multicolumn{5}{c}{Trainset} & \multicolumn{5}{c}{Testset}\\
    \cmidrule(lr){3-7}\cmidrule(lr){8-12}
    & & 1     & 3      & 5      & 7     & 10   & 1     & 3      & 5      & 7     & 10   \\
    \midrule
    \multirow{10}{*}{Qwq-32b}&Current Affairs & 70.86\% & 72.35\% & 68.54\%  & 71.08\%  & \textbf{74.83}\% & 59.57\% & 62.07\%  & 61.33\% & 57.53\% & 64.94\%\\
&Education       & 81.57\%  & 83.86\%  & 84.00\% & 85.01\%  & 85.39\% & 80.81\% & 82.98\% & 83.15\% & 85.39\%  & 82.98\% \\
&Entertainment   & 68.90\%  & \textbf{74.57}\%  & 75.53\%  & 76.60\% & 70.70\% & 71.55\% & 79.41\%& 78.74\%  & 77.52\%  & 75.20\% \\
&Fashion & 72.46\%  & 73.16\% & \textbf{85.20}\%  & 85.06\% &83.57\%  & 78.79\% & 78.48\% & 85.42\% & 79.41\% & 83.87\% \\
&Finance & 67.38\%  & 66.81\%  & \textbf{73.53}\%  & 74.04\%  & 75.21\%  & 72.13\% & 71.83\% & 74.77\% & 73.60\%  & 73.50\% \\
&Games   & 93.47\% & 95.36\% & 94.24\% & 93.75\% & 94.56\% & 90.27\%  & 93.22\% & 93.91\% & 91.07\%  & 91.89\%\\
&Home    & 86.10\% & 84.96\% & 88.33\% & 88.10\% & 88.36\% & 82.00\%  & 82.88\% & 83.33\% & 85.46\% & 85.22\% \\
&Lottery & 79.39\% & \textbf{92.47}\% & 91.69\% & 92.91\% & 95.24\% & 86.02\% & 92.86\% & 92.78\% & 95.56\% & 93.75\% \\
&Real Estate     & 86.67\% & 89.72\% & 88.43\% & 90.50\% & 87.62\%  & 81.48\%  & 88.33\% & 87.04\%  & 85.71\%  & 85.15\% \\
&Society & 62.06\% & 61.39\% & 62.33\% & 62.65\% & 66.38\% & 57.14\% & 60.76\%  & 61.22\% & 60.53\% & 62.77\% \\
&Sports  & 65.05\%  & \textbf{82.95}\% & 88.40\% & 90.69\% & 92.75\% & 72.55\% & 83.76\% & 89.08\% & 91.38\% & 92.56\% \\
&Stock   & 64.69\% & 64.32\% & 71.75\% & 72.29\% & 71.30\%  & 70.71\% & 74.07\% & 71.30\% & 75.63\%  & 73.68\%\\
&Technology      & 70.93\%& 72.83\% & 72.51\% & 73.53\% & 73.42\%  & 74.29\%  & 73.08\% & 74.75\% & 75.25\% & 73.47\% \\
&Horoscope  & 76.88\% & 78.21\% & 77.26\% & 78.21\% & 80.42\% & 80.41\% & 78.79\% & 76.40\% & 78.26\% & 80.44\% \\
&overall & 72.04\% & \textbf{76.88}\% & 79.43\% & 80.34\% & 80.55\%  & 72.81\% & 77.85\%& 79.07\% & 78.70\% & 79.44\%\\
\midrule
    \multirow{10}{*}{Qwen-max}&Current Affairs & 72.53\%  & 76.17\%  & 77.17\%  & 78.30\%  & 77.35\% & 66.67\% & 71.56\% & 73.87\% & 74.78\%  & 75.86\% \\
&Education       & 87.23\%  & 87.53\%  & 87.59\% & 87.90\% & 87.41\% & 88.14\%  & 87.93\% & 87.18\% & 85.47\% & 86.21\%\\
&Entertainment   & 75.10\%  & \textbf{80.79}\% & 80.53\%  & 79.23\% & 83.87\% & 77.03\% & 86.57\% & 81.75\%  & 84.29\%  & 87.69\%\\
&Fashion & 79.89\%  & 68.52\%  & 68.39\%  & 72.78\%  & \textbf{82.01}\%  & 77.78\%  & 66.02\% & 66.02\%  & 70.00\%  &77.48\% \\
&Finance & \textbf{52.40}\%  & \textbf{65.80}\% & 64.82\%  & 67.86\%  & 63.88\% & 52.46\%  & 67.65\%  & 68.61\%  & 68.00\%  & 72.87\% \\
&Games   & 95.14\%  & 95.70\%  & 95.43\% & 95.70\% & 95.17\% & 94.12\%  & 95.08\%  & 94.22\% & 94.22\%   & 93.33\%  \\
&Home    & 88.37\% & 82.44\%  & 80.00\% & 82.45\%  & 84.78\%  & 84.48\%  & 79.03\%  & 74.60\%  & 76.36\%  & 76.64\%\\
&Lottery & 80.70\% & 81.74\% & \textbf{91.78}\% & 93.75\% & 93.19\% & 82.00\%  & 82.00\%  & 90.91\%  & 90.91\%  & 91.89\% \\
&Real Estate     & 91.73\%  & 91.09\% & 92.16\%  & 91.22\% & 91.40\% & 89.60\%  & 89.06\%  & 88.37\%  & 87.69\%  & 89.23\% \\
&Society & 69.07\%& 71.91\% & 71.91\% & 72.04\%  & 71.95\% & 67.55\%  & 69.44\%  & 70.92\% & 65.75\% & 70.00\%  \\
&Sports  & 84.35\% & 83.48\%  & \textbf{91.55}\%  & 91.69\% & 91.00\% & 83.45\%  & 83.45\% & 88.55\%  & 88.55\%  & 89.23\% \\
&Stock   & 56.94\% & 62.30\%  & 61.16\% & 58.64\%  & \textbf{65.71}\% & 57.85\% & 64.15\%  & 63.46\%  & 53.33\% & 70.91\% \\
&Technology      & 75.58\%  & 74.05\% & 75.14\%  & 75.84\% & 76.32\% & 72.55\% & 75.73\% & 71.85\% & 74.77\% & 72.22\% \\
&Horoscope  & 76.36\%  & 80.70\%  & 79.29\%  & \textbf{86.11}\%  & \textbf{93.23}\% & 80.00\% & 87.85\%  & 82.35\% & 87.85\% & 95.73\% \\
&overall & 77.17\%  & 78.60\%  & 79.64\%  & 80.89\%  & 82.42\% & 76.46\% & 78.88\% & 78.67\% & 78.69\%  & 81.91\%\\
    \bottomrule
\end{tabular}
\end{table*}

Table~\ref{tab:Thucnews Trainset and Testset} also presents the results for the training and test sets of Qwen-max, respectively. The overall performance on Qwen-Max is inferior to that of QwQ-32b. The overall uplift on the training set is $5.25\%$. The improvement on Qwen-Max is better than that on Cluener, indicating that PromptFlow has an advantage in classification tasks. For the "Horoscope" label, we observe a significant improvement as the number of training iterations increases, rising from $76.36\%$ to $93.23\%$, demonstrating that the PromptFlow learned effectively. On the test set, the overall score improved by $5.45\%$, which is highly consistent with the results on the training set, confirming the effectiveness of our training. 

\begin{figure*}[t]
  \centering
  \begin{subfigure}[t]{0.45\linewidth}
    \includegraphics[width=\linewidth]{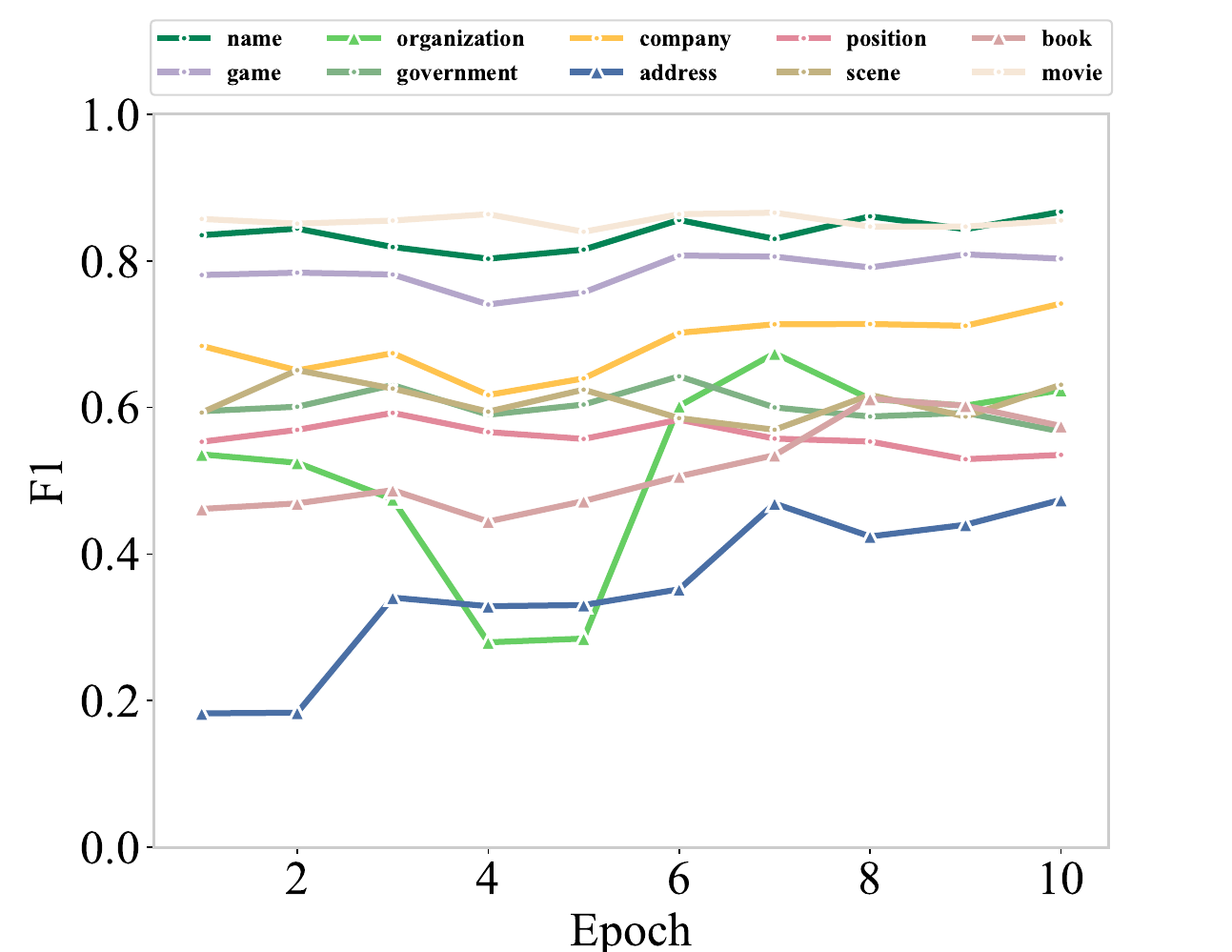}
    \caption{\small{CLS Task Module Selection}}
    \label{fig:Thucnews module development}
  \end{subfigure}
  \begin{subfigure}[t]{0.45\linewidth}
    \includegraphics[width=\linewidth]{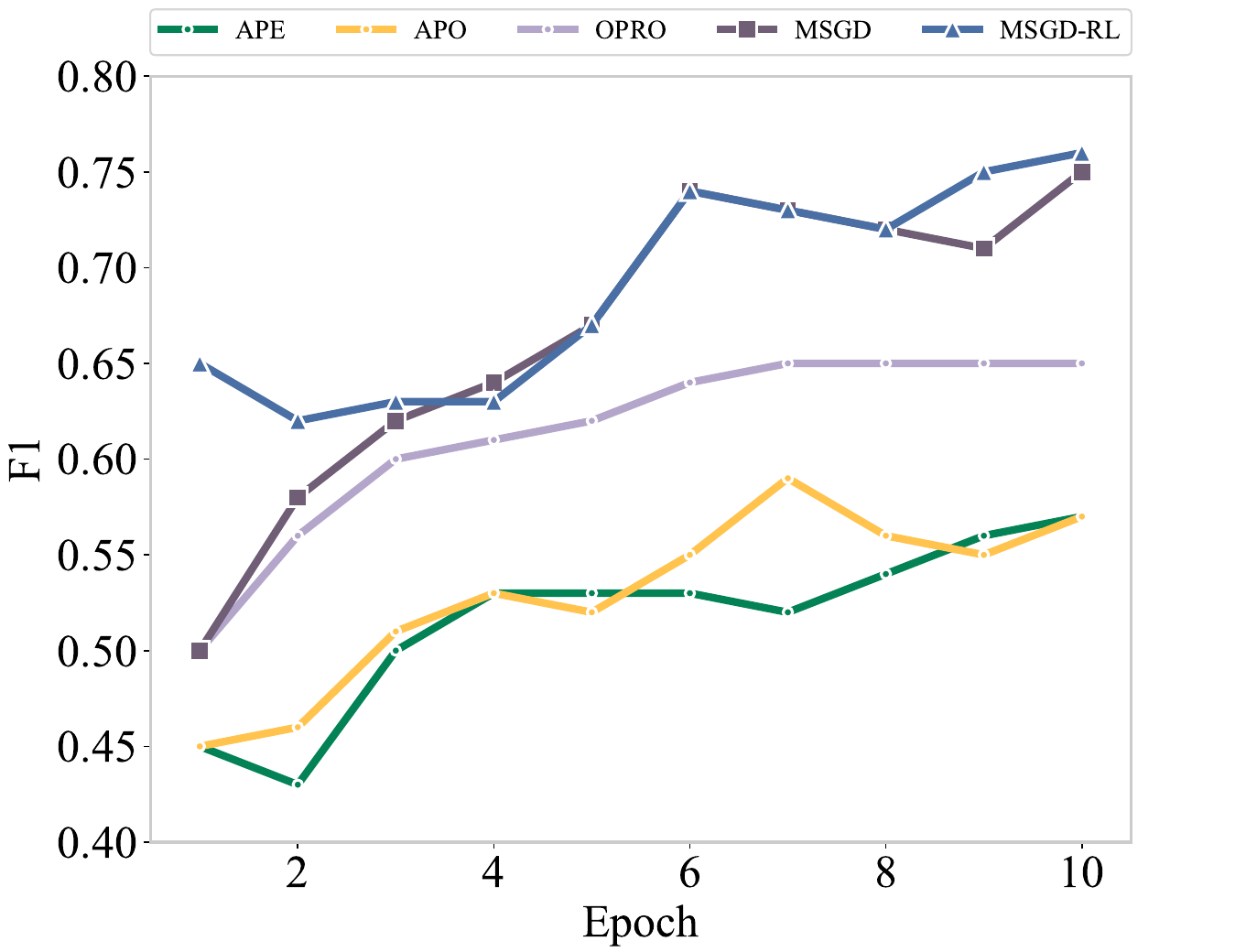}
    \caption{\small{Optimizers performance}}
    \label{fig:Optimizers performance}
  \end{subfigure}
\end{figure*}

\subsubsection{MRC Task}
%每个llm一个表，展示 列为train epoch, test acc. 行为label标签的数据
Table~\ref{tab:Squad on Trainset and Testset} presents the results for QwQ-32b and Qwen-max,respectively.
The overall performance on MRC is modest. On QwQ-32b, there is a $1.21\%$ improvement on the training set and a $1.73\%$ improvement on the test set. Compared to NER and Classification tasks, the MRC task employs English datasets and at the same time has fewer meta-prompt components that can be modified, resulting in limited overall score improvement from meta-prompt adjustments. In future work, we can explore better optimization methods.

\begin{table*}[h]
\centering
\caption{Squad Performance on Trainset and Testset}
\label{tab:Squad on Trainset and Testset}
\setlength{\tabcolsep}{3pt}
\begin{tabular}{l l *{10}{c}} % 左对齐第一列，其余列居中对齐
    \toprule
    \multirow{2}{*}{Model} & \multicolumn{5}{c}{Trainset} & \multicolumn{5}{c}{Testset}\\
    \cmidrule(lr){2-6}\cmidrule(lr){7-11}
     & 1     & 3    & 5       & 7      & 10   & 1     & 3    & 5       & 7      & 10   \\
    \midrule
    Qwq-32b & 89.42\%  & 89.88\% & 89.96\% & 90.61\%  & 90.63\% & 90.15\%  & 91.81\% & 90.77\%  & 92.58\%  & 91.88\%\\
    Qwen-max & 90.21\%  & 90.74\% & 90.52\%  & 90.74\%  & 90.19\% & 91.15\% & 91.81\% & 91.65\% & 91.81\% & 92.31\%\\
    \bottomrule
\end{tabular}
\end{table*}

\subsection{Meta-prompt Selection Analysis}\label{sec:Operator analysis}
We take the CLS Task(Thucnews) to explore the Meta-prompt selection process further. As shown in the initial prompt, each label description can be treated as a meta-prompt module, which operators can further optimize in PromptFlow. Based on Figure~\ref{fig:Thucnews module development}, it is evident that the label "address" initially receives a relatively low score but shows significant improvement over time. This indicates that enhancing the "address" label can efficiently boost the overall score. However, meta-prompt module may negatively impact others, as evidenced by the sudden drop in the "organization" score at epoch 4. PromptFlow detects the drop and subsequently achieves improvement in later iterations. The figures show fluctuations in the scores of individual labels during the development process, but ultimately result in an overall improvement by the end.

\section{Compute Cost} \label{sec:Compute cost}
Our experiments are carried out using the following resources and configurations:
\begin{itemize}[label=$\bullet$,leftmargin=*]
  \item \textbf{Hardware}:All experiments are carried out on a single machine equipped with an NVIDIA A100 GPU. The system has 64GB of RAM.
  \item \textbf{Model}:We use GPT-4, Qwen-max, QwQ-32b for prompt training.
  \item \textbf{Dataset}:We utilize the Cluener, Thucnews, and Squad datasets. Specifically, each data set contains 1,400 training samples and 600 test samples.
  \item \textbf{Multi-threading}:The number of threads is set to 100, allowing for parallel processing of data batches. 
  \item \textbf{Training Duration}:The model is trained over 10 epochs.
\end{itemize}
Details of compute cost is shown in Table~\ref{tab:Compute cost}.

\begin{table*}[ht]
\centering
\caption{Compute cost on PromptFlow on datasets}
\label{tab:Compute cost}
\begin{tabular}{@{}cccccccccc@{}}
\toprule
\multicolumn{1}{c}{\multirow{2}{*}{Dataset}} &
  \multicolumn{1}{c}{\multirow{2}{*}{Size}} &
  \multicolumn{1}{c}{\multirow{2}{*}{Iteration}} &
  \multicolumn{2}{c}{QwQ-32b} &
  \multicolumn{2}{c}{Qwen-max} &
  \multicolumn{2}{c}{GPT-4} &
   \\ \cmidrule(l){4-10} 
\multicolumn{1}{c}{} &
  \multicolumn{1}{c}{} &
  \multicolumn{1}{c}{} &
  Time(h) &
  Tokens &
  Time(h) &
  Tokens &
  Time(h) &
  Tokens &
   \\ \midrule
Thucnews & 1400 & 10 & 40.1  & 314375200  & 18.78     & 263296320 & 18.28    & 265033760 &  \\ \midrule
Cluener  & 1400 & 10 & 50.16 & 1185669760 & 13.04  & 534452800 & 14.92  & 729576640 &  \\ \midrule
Squad    & 1400 & 10 & 45.12  & 195118080  & 10.35 & 113784832 & 8.1 & 155286656 &  
 \\ \bottomrule
\end{tabular}%
\end{table*}

\end{document}